%% file: main.tex
\def\year{2021}\relax
\documentclass[runningheads]{llncs} 

\input{packages}

\usepackage{booktabs}       
\usepackage{amsfonts}       
\usepackage{nicefrac}       

\title{Enforcing Almost-Sure Reachability in POMDPs}

\author{Sebastian Junges\inst{1} \and Nils Jansen\inst{2} \and Sanjit A.\ Seshia\inst{1}}
\institute{University of California, Berkeley, USA \and Radboud University Nijmegen, The Netherlands}

\input{macros}

\begin{document}
\maketitle
\input{abstract}
\input{introduction}

\input{preliminaries}
\input{algorithm}

\input{bdd}
\begin{figure*}[t]
\centering
\frame{\includegraphics[scale=0.22]{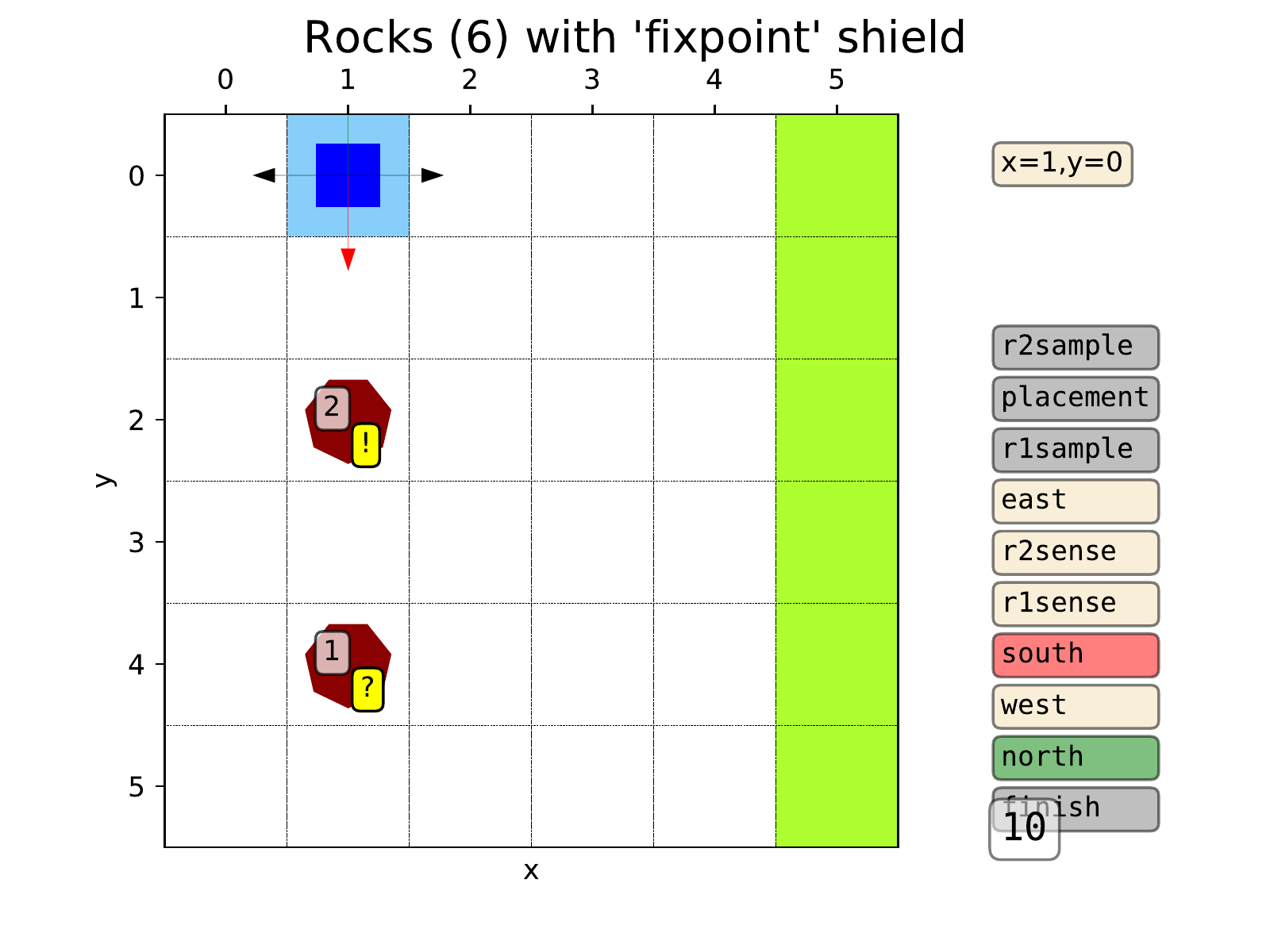}}
\hfill
\frame{\includegraphics[scale=0.22]{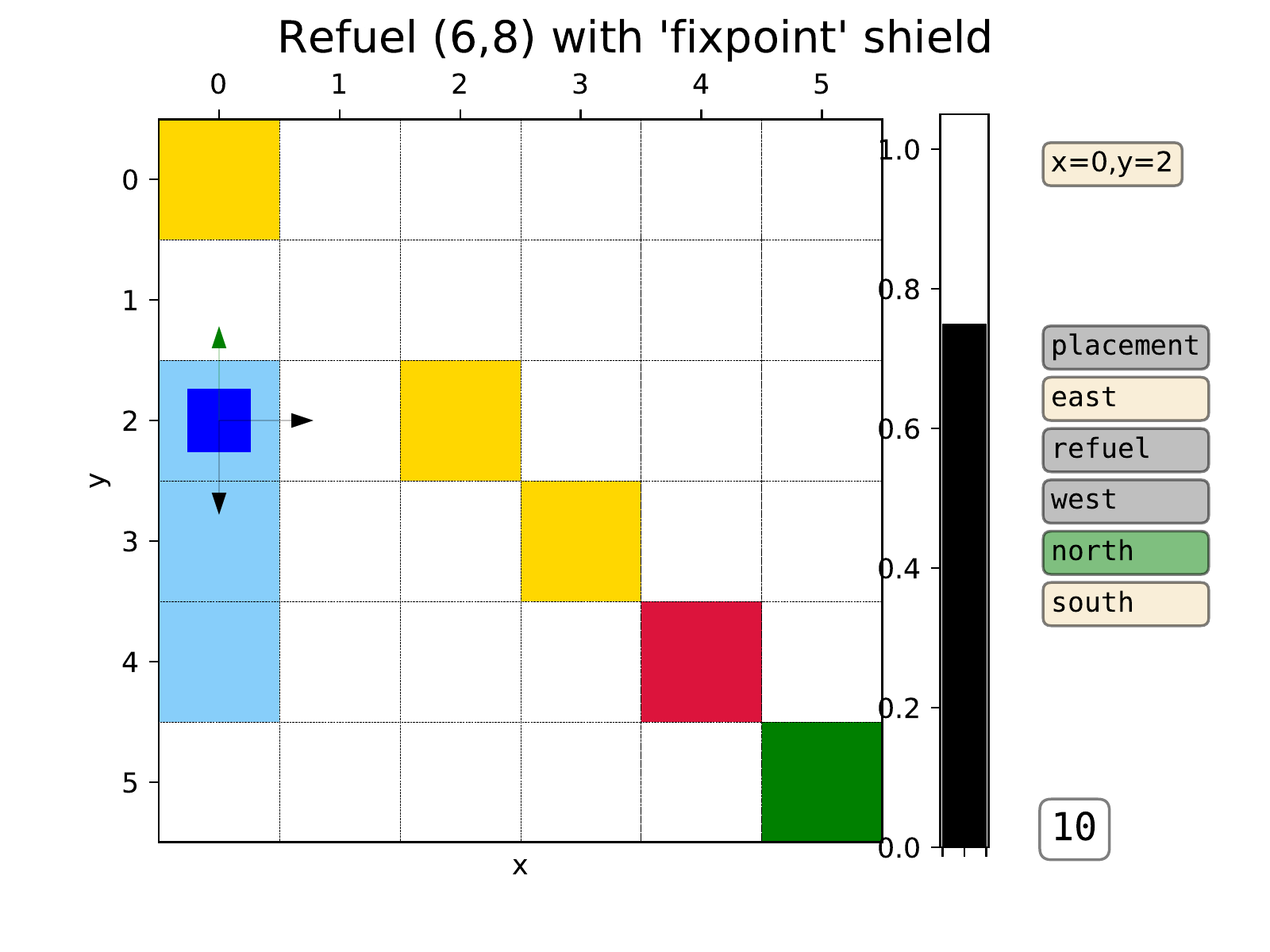}}
\hfill
\frame{\includegraphics[scale=0.22]{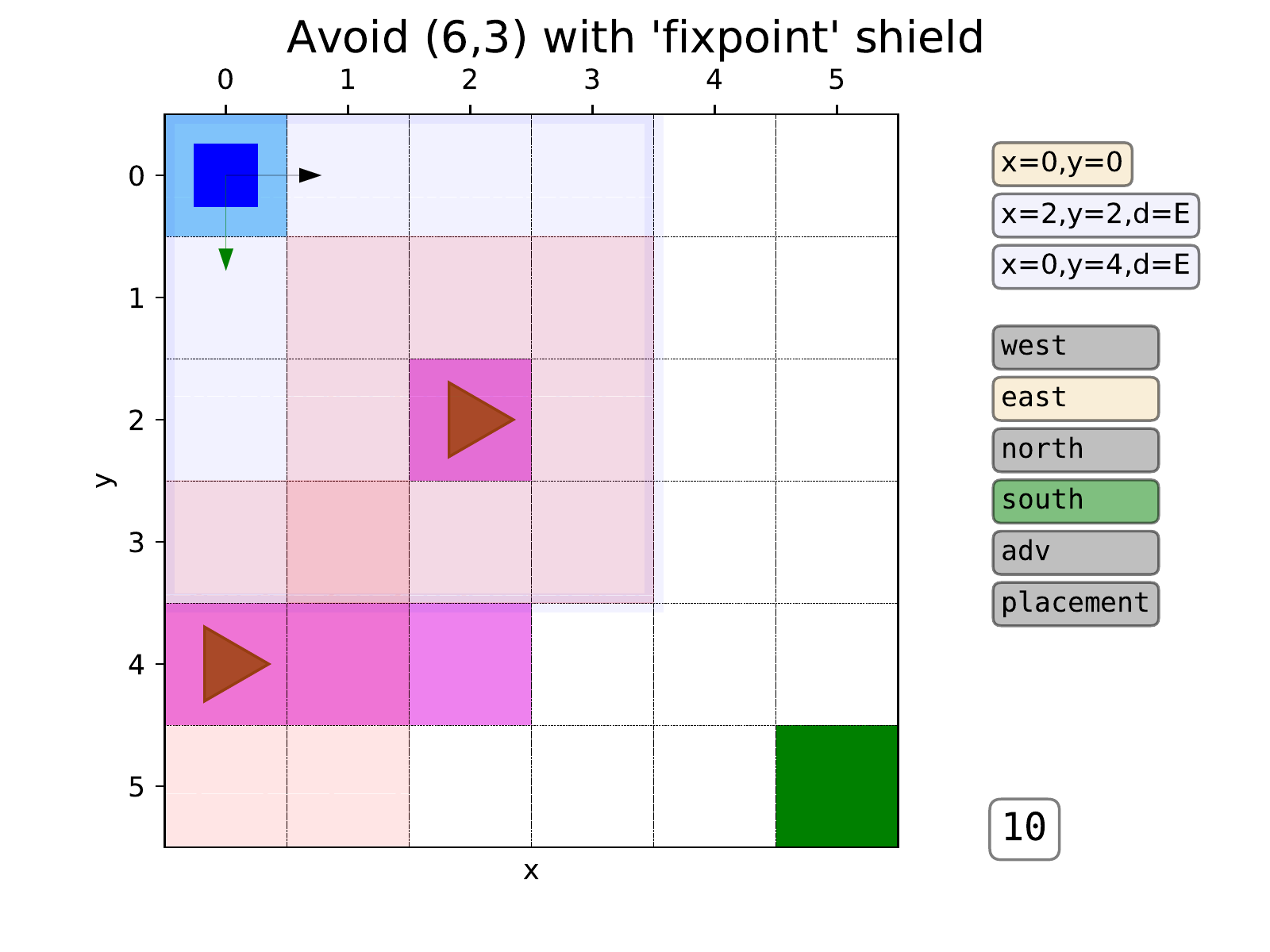}}
\caption{Video stills from simulating a shielded agent on three different benchmarks. Videos can be found online (URL omitted for blind review).}
\label{fig:screenshots}
\end{figure*}
\input{experiments}
\input{conclusion}
\clearpage
\pagebreak
\bibliographystyle{plain}
\bibliography{literature}
\newpage\appendix
\input{proof}

\end{document}

%% file: packages.tex
\usepackage{amsmath,amssymb,amsfonts}
\usepackage{mathtools}

\usepackage{amsthm}
\usepackage[usenames,dvipsnames]{color}
\usepackage[dvipsnames]{xcolor}
\usepackage{xspace}
\usepackage{microtype}
\usepackage{todonotes}
\usepackage{colonequals}
 \usepackage{booktabs}

\usepackage{amssymb, bm}

\usepackage{tabularx}
\usepackage{multirow}

\usepackage{paralist}
\usepackage{subfigure}
\usepackage{pgfplots,pgfplotstable}
\usepackage{url}
\usepackage{appendix}
\usepackage{graphicx}
\usepackage{algorithm}
\usepackage{algpseudocode}
\usepackage{listings}
\usepackage{nicefrac}
\usepackage{tikz} 
\pgfplotsset{compat=1.8}
\usepackage{mdframed}
\usepgfplotslibrary{fillbetween}
\usepgfplotslibrary{groupplots}
\usepgfplotslibrary{dateplot}
\usetikzlibrary{positioning}

\usetikzlibrary{patterns}

\usetikzlibrary{arrows,decorations.pathmorphing,positioning,fit,trees,shapes,shadows,automata,calc} 

\tikzset{outline/.style args={#1}{%
  draw=#1,thick,fill=#1!50}}

%% file: macros.tex
\definecolor{obs1}{rgb}{0.36, 0.54, 0.66}
\definecolor{obs2}{rgb}{1.0, 0.6, 0.4}
\definecolor{obs3}{rgb}{0.6, 0.47, 0.48}
\definecolor{avoid}{rgb}{0.82, 0.1, 0.26}
\definecolor{reach}{rgb}{0.0, 0.5, 0.0}

\definecolor{cheesebound}{rgb}{0.2, 0.2, 0.7}
\definecolor{sched}{rgb}{0.7, 0.2, 0.2}
\definecolor{prsched}{rgb}{0.8, 0.6, 0.2}

\tikzset{cross/.style={cross out, draw, 
         minimum size=2*(#1-\pgflinewidth), 
         inner sep=0pt, outer sep=0pt}}

\newcommand{\mdp}{\mathcal{M}}
\newcommand{\Act}{\textnormal{Act}}
\newcommand{\dinit}{\mu_\textnormal{init}}
\newcommand{\pomdp}{\mathcal{P}}
\newcommand{\post}[2]{\textnormal{post}_{#1}(#2)}
\newcommand{\enact}[1]{\textnormal{EnAct}(#1)}
\newcommand{\obsfun}{\textnormal{obs}}

\newcommand{\lifted}[1]{#1_\belsups}

\newcommand{\highlight}[1]{\textcolor{blue!80!red}{#1}}
\newcommand{\timecol}{red!70!black}
\newcommand{\runtime}[1]{\textcolor{\timecol}{#1}}

\newcommand{\prob}{\mathbf{P}}
\newcommand{\Obs}{\Omega}
\newcommand{\obs}{z}
\newcommand{\probability}{\mathit{Pr}}

\newcommand{\beliefs}{\Distr(S)}

\newcommand{\belief}{\mathfrak{b}}
\newcommand{\belsup}{b}
\newcommand{\belsups}{B}
\newcommand{\sched}{\sigma}
\newcommand{\avoid}{\textsl{AVOID}}
\newcommand{\reach}{\textsl{REACH}}
\newcommand{\questionmark}{{S_{?}}}

\newcommand{\psched}{\nu}

\newcommand{\trace}{\tau}

\newcommand{\support}[1]{\textsl{supp}(#1)}

\newcommand{\winning}[1]{\ensuremath{W_{#1}}}

\newcommand{\maxrank}{\ensuremath{k}}

\newcommand{\eg}{e.g.\@\xspace}

\newcommand{\prism}{\textrm{PRISM}\xspace}
\newcommand{\storm}{\textsc{Storm}\xspace}

\newcommand{\Distr}{\mathit{Distr}}

\newcommand{\distFunc}{\mu}

\newcommand{\supp}{\mathit{supp}}
\newcommand{\bigland}{\ensuremath{\bigwedge}}
\newcommand{\biglor}{\ensuremath{\bigvee}}
\newcommand{\last}[1]{\ensuremath{\mathsf{last}(#1)}}

\newcommand{\act}{\ensuremath{\alpha}}
\newcommand{\beliefsupportmdp}{\pomdp_\belsups}

\newcommand{\encact}[2]{\ensuremath{A_{#1,#2}}}
\newcommand{\encreached}[1]{\ensuremath{C_{#1}}}
\newcommand{\encreach}[2]{\ensuremath{P_{{#1,#2}}}}
\newcommand{\encctd}[1]{\ensuremath{D}_{#1}}
\newcommand{\encswitch}[1]{\ensuremath{F}_{#1}}
\newcommand{\encupdated}[1]{\ensuremath{U}_{#1}}
\newcommand{\encprevpol}[1]{\ensuremath{I}_{#1}}
\newcommand{\encrank}[1]{\ensuremath{R}_{#1}}

\newcommand{\winningdata}{\mathsf{Win}}

\newcommand{\encmemless}[4]{\ensuremath{\Psi_{#1}^{#2}(#3,#4)}}
\newcommand{\enciterative}[3]{\ensuremath{\Phi_{#1}^{#2}(#3)}}           
           
           \newcommand{\zthree}{\textsc{z3}}
           \newcommand{\oom}[1]{\ensuremath{\textsc{e}\mathbf{#1}}}
           
\algdef{SE}[DOWHILE]{Do}{doWhile}{\algorithmicdo}[1]{\algorithmicwhile\ #1}%

\newcommand{\transitions}{\prob}
\newcommand{\state}{s}
\newcommand{\states}{S}
\newcommand{\action}{\act}
\newcommand{\observation}{\obs}
\newcommand{\obsof}[1]{\obsfun(#1)}
\newcommand{\iverson}[1]{[#1]}
\newcommand{\beliefstate}{\belief}
\newcommand{\nextbelief}[3]{\textsl{update}(#1|#2,#3)}
\newcommand{\beliefmdp}[1]{\textsl{BelMDP}(#1)}
\newcommand{\beliefstates}{\mathcal{B}}
\newcommand{\belieftransitions}{\prob_\beliefstates}

%% file: abstract.tex
\begin{abstract}
Partially-Observable Markov Decision Processes (POMDPs) are a well-known stochastic model for sequential decision making under limited information.
We consider the EXPTIME-hard problem of synthesising policies that almost-surely reach some goal state without ever visiting a bad state. 
In particular, we are interested in computing the winning region, that is, the set of system configurations from which a policy exists that satisfies the reachability specification.
A direct application of such a winning region is the safe exploration of POMDPs by, for instance, restricting the behavior of a reinforcement learning agent to the region. We present two algorithms: 
A novel SAT-based iterative approach and a decision-diagram based alternative. 
The empirical evaluation demonstrates the feasibility and efficacy of the approaches.
\end{abstract}

%% file: introduction.tex
\section{Introduction}
\label{sec:introduction}
Partially observable Markov decision processes (POMDPs) constitute the standard model for agents acting under partial information in uncertain environments~\cite{kaelbling1998planning,thrun2005probabilistic}.
A common problem is to find a policy for the agent that maximizes a reward objective~\cite{MadaniHC99}.
This problem is undecidable, yet, well-established approximate~\cite{hauskrecht2000value}, point-based~\cite{pineau2003point}, or Monte-Carlo-based~\cite{silver2010monte} methods exist.
In safety-critical domains, however, one seeks a \emph{safe} policy that exhibits strict behavioral guarantees, for instance in the form of temporal logic constraints~\cite{Pnueli77}.
The aforementioned methods are not suitable to deliver provably safe policies.
In contrast, we employ almost-sure reach-avoid specifications, where the probability to reach a set of \emph{avoid} states is zero, and the probability to \emph{reach} a set of goal states is one.
Our \textbf{Challenge~1} is to compute a policy that adheres
to such specifications.
Furthermore, we aim to ensure the \emph{safe exploration of a POMDP}, with safe reinforcement learning~\cite{DBLP:journals/jmlr/GarciaF15} as direct application.
\textbf{Challenge~2} is then to compute a large set of safe policies for the agent to choose from at any state of the POMDP.
Such sets of policies are referred to as \emph{permissive policies}~\cite{draeger-et-al-tacas-2014,junges-et-al-tacas-2016}.


\paragraph{POMDP Almost-Sure Reachability Verification.}
Let us remark that in POMDPs, we cannot directly observe in which state we are, but we are in general able to track a \emph{belief}, i.e., a distribution over states that describes where in the POMDP we may be.
The belief allows us to formulate the following \textbf{verification task}:
\begin{center}
\vspace{-0.3em}
\fcolorbox{black}{orange!20!white!95!black}{
\parbox[l]{0.95\columnwidth}{
For a POMDP, sets of target and avoid states, and a belief, does a policy exist such that we reach the target states without ever visiting a bad state?	
}}
\vspace{-0.3em}
\end{center}%
%
%
%
\noindent The underlying EXPTIME-complete problem requires---in general---policies with access to memory of exponential size in the number of states~\cite{DBLP:journals/jacm/BaierGB12,DBLP:conf/mfcs/ChatterjeeDH10}.
For safe exploration and, e.g., to support nested temporal properties, the ability to solve this problem \emph{for each belief in the POMDP} is essential.

We base our approaches on the concept of a \emph{winning region}, also referred to as controllable or attractor regions.
Such regions are sets of \emph{winning beliefs} from which a policy exists that guarantees to satisfy an almost-sure specification.
The verification task relates three concrete problems which we tackle in this paper: (1) \emph{Decide} whether a belief is winning, (2) \emph{compute} the \emph{maximal} winning region, and (3) \emph{compute} a \emph{large} yet not necessarily maximal winning region.
We now outline our two approaches. 
First, we directly exploit model checking for MDPs~\cite{BK08} using belief abstractions.
The second and much faster approach iteratively exploits \emph{Boolean satisfiability solving} (SAT)~\cite{DBLP:series/faia/2009-185}.
Finally, we define a scheme to enable safe reinforcement learning~\cite{DBLP:journals/jmlr/GarciaF15} for POMDPS, referred to as \emph{shielding}~\cite{shield_rl,jansen-et-al-shield}.
\paragraph{MDP model checking.}
A prominent approach describes the semantics of a POMDP via an (infinite) belief MDP whose state space is the set of beliefs in the POMDP~\cite{MadaniHC99}.
For almost-sure specifications, it is sufficient to consider \emph{belief-supports} rather than beliefs.
In particular, two beliefs with the same support are either both in a winning region or not~\cite{DBLP:journals/lmcs/RaskinCDH07}.
We abstract a belief MDP into a finite belief-support MDP, whose states are the support of beliefs.
The (maximal) winning region are (all) states of the belief-support MDP from which one can almost surely reach a belief support that contains a goal state without visiting belief support states that contain an  avoid state.

To find a winning region in the POMDP, we thus just have to solve  almost-sure reachability in this finite MDP.
The number of belief supports, however, is exponentially large in the number of POMDP states, threatening the efficient application of explicit state verification approaches.
Symbolic state space representations are a natural option to mitigate this problem~\cite{DBLP:conf/aips/BertoliCP06}.
We construct a symbolic description of the belief support MDP and apply state-of-the-art symbolic model checking.
Our experiments, however, show that this approach (referred to as \emph{MDP Model Checking}) does in general fail to alleviate the exponential blow-up.

\paragraph{Incremental SAT solving.}
While the belief support model exploits the structure of the structure of the belief support MDP by using a symbolic state space representation, it does not exploit elementary properties of the structure of winning regions.
To overcome the scalability challenge, we aim to exploit information from the original POMDP, rather than working purely on the belief-support MDP. 
In a nutshell, our approach computes the winning regions in a backward fashion by \emph{optimistically} searching policies without memory on the POMDP level. Concretely, starting from the belief support states that shall be reached almost-surely, further states are added to the winning region if we quickly can find a policy that reaches these states without visiting those that are to avoid.
We search for these policies by incrementaly employing an encoding based on 
SAT solving.
This symbolic encoding avoids an expensive construction of the belief support MDP.
The computed winning region directly translates to sufficient constraints on the set of safe policies, i.e., each policy satisfying these constraints satisfies, by construction, the specification.
The key idea is to successively add short-cuts corresponding to already known safe policies. 
These changes to the structure of the POMDP are performed implicitly on the SAT encoding.
The resulting scalable method is sound, but not complete by itself. However, it can be rendered complete by trading off a certain portion of the scalability; the intuition is that 
one has to eventually search for policies with larger amounts of memory.

\paragraph{Shielding. }
An agent that stays within a winning region is guaranteed to adhere to the specification.
In particular, we \emph{shield} (or \emph{mask}) any action of the agent that may lead out of the winning region~\cite{DBLP:journals/tsmc/NamA10,DBLP:conf/mesas/PeckaS14,DBLP:conf/cdc/AkametaluKFZGT14}.
We stress that the shape of the winning region is independent of the transition probabilities or rewards in the POMDP.
This independence means that the only prior knowledge we need to assume is the topology, that is, the graph of the POMDP. 
A pre-computation of the winning region thus yields a shield and allows us to restrict an agent to safely explore environments, which is the essential requirement for safe reinforcement learning~\cite{DBLP:journals/jmlr/GarciaF15,DBLP:conf/aaai/FultonP18} of POMDPs.
The shield can be used with any RL agent~\cite{shield_rl}.

\paragraph{Comparison with the state-of-the-art.}
Similar to our approach,~\cite{DBLP:conf/aaai/ChatterjeeCD16} solves almost-sure specifications using SAT.
Intuitively, the aim is to find a so-called \emph{simple policy} that is Markovian (aka memoryless).
Such a policy may not exist, yet, the method can be applied to a POMDP that has an extended state space to account for finite memory~\cite{meuleau1999solving,junges2018finite}.
There are three shortcomings that our incremental SAT approach overcomes.
First, one needs to pre-define the memory a policy has at its disposal, as well as a fixed lookahead on the exploration of the POMDP. 
Our encoding does not require to fix these hyperparameter a priori.
Second, the approach is only feasible if small memory  bounds suffice. 
Our approach scales to models that require policies with larger memory bounds.
Third, the approach finds a single simple policy starting from a pre-defined initial state. 
Instead, we find a large winning region.
For safe exploration, this means that we may exclude many policies and never explore important parts of the system, harming the final performance of the agent.
Shielding MDPs is not new~\cite{DBLP:conf/tacas/BloemKKW15,shield_rl,jansen-et-al-shield,DBLP:journals/corr/abs-2006-16688}.
However, those methods do neither take partial observability into account, nor can they guarantee reaching desirable states.
Nam and Alur~\cite{DBLP:journals/tsmc/NamA10} cover partial observability and reachability, but do not account for stochastic uncertainty.

\paragraph{Experiments.}
To showcase the feasibility of our method, we adopted a number of typical POMDP environments. 
We demonstrate that our method scales better than the state of the art.
We evaluate the shield by letting an agent explore the POMDP environment according to the permissive policy, thereby enforcing the satisfaction of the almost-sure specification. We visualize the resulting behavior of the agent in those environments with a set of videos.

\paragraph{Contributions.}
Our paper makes four key contributions:
(1)~We present an incremental SAT-based approach to compute policies that satisfy almost-sure properties.
The method scales to POMDPs whose belief-support states count billions; 
(2)~The novel approach is able to find large winning regions that yield permissive policies.
%
(3)~We implement a straightforward approach that constructs the belief-support symbolically using state-of-the-art model checking. 
We show that its completeness comes at the cost of limited scalability.
%
(4)~We construct a shield for almost-sure specifications on POMDPs which enforces at runtime that \emph{no unsafe states are visited} and that, under mild assumptions, \emph{the agent almost-surely reaches the set of desirable states}. 


 \paragraph{Further related work.}
Chatterjee et al. compute winning regions for minimizing a reward objective via an explicit state representation~\cite{DBLP:journals/ai/ChatterjeeCGK16}, or consider almost-sure reachability using an explicit  state space~\cite{DBLP:conf/icra/ChatterjeeCGK15,DBLP:conf/hybrid/SvorenovaCLECCB15}.
The problem of determining any winning policy can be cast as a strong cyclic planning problem, proposed earlier with decision diagrams~\cite{DBLP:conf/aips/BertoliCP06}. Indeed, our BDD-based implementation on the belief-support MDP can be seen as a reimplementation of that approach. 

Quantitative variants of reach-avoid specifications have gained attention in, e.g., \cite{DBLP:journals/rts/Norman0Z17,DBLP:conf/ijcai/HorakBC18,DBLP:conf/atva/BorkJKQ20}.
Other approaches restrict themselves to simple policies~\cite{DBLP:conf/nips/PoupartB03,DBLP:journals/aamas/AmatoBZ10,junges2018finite,DBLP:conf/nfm/Winterer00020}.
The verification of recurrent neural network policies for POMDPs against quantitative reachability specifications is introduced in~\cite{DBLP:conf/ijcai/CarrJT20,DBLP:conf/ijcai/Carr0WS0T19}.
Wang et al.~\cite{DBLP:conf/atal/WangCK18} use an iterative Satisfiability Modulo Theories (SMT)~\cite{barrett-smtbookch09} approach for quantitative finite-horizon specifications, which requires computing beliefs. 
Various general POMDP approaches exist, e.g.,~\cite{hauskrecht2000value,ShaniPK13,DBLP:conf/aaai/WalravenS17,silver2010monte,DBLP:conf/nips/JaakkolaSJ94,wierstra2007solving,hausknecht2015deep}.
The underlying approaches depend on discounted reward maximization and can satisfy almost-sure specifications with high reliability. 
However, enforcing probabilities that are close to $0$ or $1$ requires a discount factor close to $1$, drastically reducing the scalability of such approaches~\cite{DBLP:conf/ijcai/HorakBC18}.
Moreover, probabilities in the underlying POMDP need to be precisely given, which is not always realistic~\cite{burns2007sampling}.

Another line of work (for example~\cite{DBLP:conf/nips/TurchettaB019}) uses an idea similar to winning regions with uncertain specifications, but in a fully observable setting.
Finally, complementary to shielding, there are approaches that guide reinforcement learning (with full observability) via temporal logic constraints~\cite{DBLP:conf/atal/HasanbeigAK20,DBLP:conf/tacas/HahnPSSTW19}.

%% file: preliminaries.tex
\section{Preliminaries and Formal Problem}
We briefly introduce POMDPs and their semantics in terms of belief MDPs, before formalising and studying the problem variants outlined in the introduction.
We present belief-support MDPs as a finite abstraction of infinite belief MDPs.

The support of a discrete probability distribution $\distFunc$ over $X$ is denoted $\support{\distFunc}=\{x\in X\mid \distFunc(x)>0\}$, with $\Distr(X)$ the set of all distributions. 
\begin{definition}[MDP] 
	A \emph{Markov decision process} (MDP) is a tuple $\mdp = \langle S, \Act, \dinit, \prob \rangle$ with a set $S$ of states, an initial distribution $\dinit \in \Distr(S)$, a finite set $\Act$ of actions, and a  transition function
	$\prob \colon S \times \Act \to \Distr(S)$.
\end{definition}
Let $\post{s}{\act} = \support{\prob(s,\act)}$ denote the states that may be the successors of the state $s \in S$ for action $\act \in \Act$ under the distribution $\prob(s,\act)$. 
If $\post{s}{\act}=\{s\}$ for all actions $\act$, $s$ is called \emph{absorbing}.
\begin{definition}[POMDP]
	A \emph{partially observable MDP} (POMDP) is a tuple $\pomdp = \langle \mdp, \Obs, \obsfun \rangle$ with $\mdp = \langle S, \Act, \dinit, \prob \rangle$ the underlying MDP with finite $S$, $\Obs$ a finite set of observations, and $\obsfun \colon S \to \Obs$ an observation function.
We assume that there is a unique initial observation, i.e., that $|\{ \obsfun(s) \mid s \in \supp(\dinit)\}| = 1$.
\end{definition}
More general observation functions $\obsfun \colon S \to \Distr(\Obs)$ are possible via a (polynomial) reduction~\cite{DBLP:journals/ai/ChatterjeeCGK16}.
%
A path through an MDP is a sequence $\pi$, 
$\pi = (s_0,\act_0)(s_1,\act_1) \hdots s_n$
 of states and actions.
such that $s_{i+1} \in \post{s_i}{\act_i}$ for $\act_{i} \in \Act$ and $0\leq i < n$.
The observation function $\obsfun$ applied to a path yields an observation(-action) sequence $\obsfun(\pi)$
of observations and actions.%

For modeling flexibility, we allow actions to be unavailable in a state (e.g., opening doors is only available when at a door), and it turned out to be crucial to handle this explicitly in the following algorithms.
Technically, the transition function is a partial function, and the enabled actions are a set $\enact{s} = \{ \act \in \Act \mid \post{s}{\act} \neq \emptyset \}$. To ease the presentation, we assume that states $s,s'$ with the same observation share a set of enabled actions $\enact{s}=\enact{s'}$.
\begin{definition}[Policy]
A policy  $\sched \colon (S\times\Act)^*\times S \rightarrow \Distr(\Act)$ maps a path $\pi$ to a distribution over  actions.
A policy is \emph{observation-based}, if for each two paths $\pi$, $\pi'$ it holds that $\obsfun(\pi) = \obsfun(\pi') \Rightarrow \sched(\pi) = \sched(\pi').$
A policy is \emph{memoryless}, if for each $\pi$, $\pi'$ it holds that $\last{\pi} = \last{\pi'} \Rightarrow \sched(\pi) = \sched(\pi')$.
A policy is \emph{deterministic}, if for each $\pi$, $\sched(\pi)$ is a Dirac distribution, i.e., if $|\supp(\sched(\pi))| = 1$.
\end{definition}
\noindent Policies resolves nondeterminism and partial observability by turning a (PO)MDP into the \emph{induced} infinite discrete-time Markov chain
whose states are the finite paths of the (PO)MDP. 
Probability measures are defined on this Markov chain.
%
%

For POMDPs, a \emph{belief} describes the probability of being in certain state based on an observation sequence.
Formally, a belief $\belief$ is a distribution $\belief\in\Distr(S)$ over the states.
A state $s$ with {positive belief} $\belief(s)>0$ is in the \emph{belief support}, $s\in\support{b}$.
Let $\probability_\belief^\sched(S')$ denote the probability to reach a set $S'\subseteq S$ of states from belief $\belief$ under the policy $\sched$. More precisely, $\probability_\belief^\sched(S')$ denotes the probability of all paths that reach $S'$ from $\belief$ when nondeterminism is resolved by $\sched$.

The policy synthesis problem usually consists in finding a policy that satisfies a certain specification for a POMDP. 
We consider \emph{reach-avoid} specifications, a subclass of indefinite horizon properties~\cite{Put94}.
	For a POMDP $\pomdp$ with states $S$, such a specification is $\varphi=\langle \reach, \avoid \rangle \subseteq S \times S$. 
We assume that states in $\avoid$ and in $\reach$ are (made) absorbing and $\reach\cap\avoid=\emptyset$.


\begin{definition}[Winning]
	A policy $\sched$ is \emph{winning} for $\varphi$ from belief $\belief$ in (PO)MDP $\pomdp$ iff $\probability_\belief^\sched(\avoid)=0$ and $\probability_\belief^\sched(\reach)=1$, i.e., if it reaches $\avoid$ with probability zero and $\reach$ with probability one (almost-surely) when $\belief$ is the initial state.
	%
	Belief $\belief$ is \emph{winning} for $\varphi$ in $\pomdp$ if there exists a winning policy from~$\belief$. 
\end{definition}
\noindent We omit $\pomdp$ and $\varphi$ whenever it is clear from the context and simply call $\belief$ winning.

\begin{center}
\fcolorbox{black}{orange!20!white!95!black}{
\parbox[l]{0.95\columnwidth}{
\textbf{Problem~1:} Given a POMDP, a belief $\belief$, and a specification $\varphi$, decide whether $\belief$ is winning and find a policy $\sched$ that is winning from $\belief$.
}
}
\end{center}
The problem is EXPTIME-complete~\cite{DBLP:conf/mfcs/ChatterjeeDH10}. Contrary to MDPs, it is not sufficient to consider memoryless policies. 

Model checking queries for POMDPs often rely on the analysis of the \emph{belief MDP}. Indeed, we may analyse this generally infinite  model.
Let us first recap a formal definition of the belief MDP, using the presentation from \cite{DBLP:conf/atva/BorkJKQ20}.
In the following, let $\transitions(\state,\action,\observation) \colonequals \sum_{\state'\in\states} \iverson{\obsof{\state'} {=} \observation} \cdot \transitions(\state,\action,\state')$ denote the probability\footnote{We use Iverson brackets: $\iverson{x}=1$ if $x$ holds and $0$ otherwise.} to move to (a state with) observation $\observation$ from state $\state$ using action $\action$.
Then,
$\transitions(\beliefstate, \action, \observation) \colonequals \sum_{\state \in \states} \beliefstate(\state) \cdot  \transitions(\state,\action,\observation)$ is the probability to observe $\observation$ after taking $\action$ in $\beliefstate$.
We define the \emph{belief obtained by taking $\action$ from $\beliefstate$, conditioned on observing~$\observation$}:
\begin{align} \nextbelief{\beliefstate}{\action}{\observation}(\state') \colonequals  \frac{\iverson{\obsof{\state'}{=}\observation} \cdot \sum_{\state \in \states} \beliefstate(\state) \cdot \transitions(\state, \action,\state')}{\transitions(\beliefstate, \action, \observation)}. 
\label{eq:update}\end{align}

\begin{definition}[Belief MDP]
\label{def:beliefmdp}
	The \emph{belief MDP} of POMDP $\pomdp = \langle \mdp, \Obs, \obsfun \rangle$ where 
	$\mdp = \langle S, \Act, \dinit, \prob \rangle$ 
	is the MDP $\beliefmdp{\pomdp} \colonequals \langle \beliefstates, \Act, \belieftransitions, \dinit \rangle$ with
	$\beliefstates = \beliefs$, and transition function $\belieftransitions$ 
	given by 
		\[ \belieftransitions(\beliefstate,\action,\beliefstate') \colonequals \begin{cases} \belieftransitions(\beliefstate, \action,  \obsof{\beliefstate'}) & \text{if } \beliefstate' = \nextbelief{\beliefstate}{\action}{\obsof{\beliefstate'}}, \\
 0 & \text{otherwise.}	
 \end{cases}
 \]
\end{definition}
Due to \eqref{eq:update} and the unique initial observation, we may restrict the beliefs 
to $\beliefstates = \bigcup_{\obs \in \Obs} \Distr(\{ s \mid \obsfun(s) = \obs \})$, that is, each belief state has a unique associated observation. We can lift specifications to belief MDPs: \emph{Avoid-beliefs} are the set of beliefs $\beliefstate$ such that $\supp(\beliefstate) \cap \avoid \neq \emptyset$, and \emph{reach-beliefs} are the set of beliefs $\beliefstate$ such that $\supp(\beliefstate) \subseteq \reach$.

Towards obtaining a finite abstraction, the main algorithmic idea is the following. 
For the qualitative reach-avoid specifications we consider, the belief probabilities are irrelevant---\emph{only the belief support is important}~\cite{DBLP:journals/lmcs/RaskinCDH07}. 
\begin{lemma}
For winning belief $\belief$, belief $\belief'$ with $\supp(\belief)=\supp(\belief')$ is winning.	
\end{lemma}

\noindent Consequently, we can abstract the belief MDP into a finite belief support MDP.
\begin{definition}[Belief MDP]
For a POMDP $\pomdp = \langle \mdp, \Obs, \obsfun \rangle$ with $\mdp = \langle S, \Act, \dinit, \prob \rangle$, the finite state space of a \emph{belief-support MDP} $\beliefsupportmdp$ is $\belsups = \bigl\{ \belsup \subseteq S\mid\forall s,s'\in \belsup \colon \obsfun(s)=\obsfun(s')\bigr\}$ where each state is the support of a belief state. 
Action $\act$ in state $\belsup$ leads (with an irrelevant positive probability $p>0$) to a state $\belsup'$, if there is an $s \in \belsup$ and $s' \in \belsup'$ such that $s' \in \post{s}{\act}$.
\end{definition}
Thus, transitions between states within $\belsup$ and $\belsup'$ are mimicked in the POMDP.
Equivalently, the following clarifies the belief-support MDP as an abstraction of the belief MDP: there are transitions with action $\act$ between $\belsup$ and $\belsup'$, if there exists beliefs $\belief,\belief'$ with $\supp(\belief) = \belsup$ and $\supp(\belief') = \belsup'$, such that $\belief' \in \post{\belief}{\act}$. 

\noindent We lift the specification as before:
\begin{definition}[Lifted specification]
	For $\varphi = \langle \avoid, \reach \rangle$, we define $\lifted{\varphi} = \langle \lifted{\avoid}, \lifted{\reach} \rangle$ with $\lifted{\avoid} = \{ \belsup \mid \belsup \cap \avoid \neq \emptyset \}$, and $\lifted{\reach} = \{ \belsup \mid \belsup \subseteq \reach \}$.
\end{definition}

We obtain the following lemma, which follows from the fact that almost-sure reachability is a graph property\footnote{Although the probabilities are not relevant to compute almost-sure reachability, it is important to notice that almost-sure reachability is different from sure-reachability~\cite{BK08}: For almost-sure reachability, there can be an infinite path that never reaches the target, as long as the probability mass over all those paths is $0$. Almost-sure reachability can, however, be expressed as sure-reachability in a particular game-setting~\cite{DBLP:journals/lmcs/RaskinCDH07}}. 
\begin{lemma}
\label{lem:belsupmdpeq}
If belief $\beliefstate$ is winning in the POMDP $\pomdp$ for $\varphi$, then the support $\supp(\beliefstate)$ is winning in the belief-support MDP $\beliefsupportmdp$ for $\lifted{\varphi}$.
\end{lemma}
\noindent
Lemma~\ref{lem:belsupmdpeq} yields an equivalent reformulation of Problem~1 for belief supports:
\begin{center}
\fcolorbox{black}{white}{
\parbox[l]{0.95\columnwidth}{
\textbf{Problem~1 (equivalent):} Given a POMDP $\pomdp$, belief $\belief$, and specification $\varphi$, decide whether $\supp(\belief)$ is winning for $\lifted{\varphi}$ in the belief-support MDP $\beliefsupportmdp$.
}
}
\end{center}

\section{Winning Regions and Shields}
In this section, we make key observations that motivate framing the remainder of the paper in terms of winning regions. 

\subsection{Winning regions}
An important consequence of Lemma~\ref{lem:belsupmdpeq} and the reformulation of Problem~1 to the belief-support MDP
is that the initial distribution of the POMDP is no longer relevant.
Winning policies for individual beliefs may be composed to a policy that is winning for all of these beliefs, using the individual action choices.
\begin{lemma}\label{lemma:winning_support_composable_policies}
If the policies $\sched$ and $\sched'$ are winning for the belief supports $\belsup$ and $\belsup'$, respectively, then there exists a policy $\sched''$ that is winning for both $\belsup$ and $\belsup'$.
\end{lemma}
\noindent While this statement may seem trivial on the MDP (or equivalently on beliefs), we notice that it does not hold for POMDP states. 
As a natural consequence, we are able to consider winning beliefs without referring to a specific policy.  
\begin{definition}[Winning region]
Let $\sched$ be a policy.
	A set $\winning{\varphi}^{\sched} \subseteq \belsups$ of belief supports is a \emph{winning region for $\varphi$ and $\sched$}, if $\sched$ is winning from each $\belsup \in \winning{\varphi}^{\sched}$. 
	A set $\winning{\varphi} \subseteq \belsups$ is a winning region for $\varphi$, if every $\belsup\in\winning{\varphi}$ is winning. The region containing all winning beliefs is the \emph{maximal winning region}.
\end{definition}
\noindent Observe that the maximal winning region in MDPs  exists for qualitative reachability, but not for quantitative reachability, which we do not consider here.
\begin{center}
\fcolorbox{black}{orange!20!white!95!black}{
\parbox[l]{0.95\columnwidth}{
\textbf{Problem~2:} Given a POMDP $\pomdp$ and a specification $\varphi$, find the maximal winning region $\winning{\varphi}$.  	
}
}
\end{center}
Using this definition of winning regions, we are able to reformulate \textbf{Problem 1} by asking whether the support of some belief $\belief$ is in the winning region.

Part of \textbf{Problem 1} was to compute a winning policy. Below, we study the connection between the winning region and winning policies. 
We are interested in subsets of the maximal winning region that exhibit two properties:
\begin{definition}[Deadlock-free]
	A set $W$ of belief-supports $W\subseteq \belsups$ is \emph{deadlock-free}, if for every $\belsup \in W$, an action $\act \in \enact{\belsup}$ exists such that $\post{\belsup}{\act} \subseteq W$. 
\end{definition}
\begin{definition}[Productive]
   A set of belief supports $W \subseteq \belsups$ is \emph{productive} (towards a set $\lifted{\reach}$), if from every $\belsup \in W$, a (finite) path $\pi = \belsup_0\act_1\belsup_1\hdots\belsup_n$ from $\belsup_0$ to $\belsup_n \in \lifted{\reach}$ with $\belsup_i \in W$ and $\post{\belsup_i}{\act} \subseteq W$ for all $1 \leq i \leq n$.
\end{definition}
\noindent Every productive region is deadlock-free, as $\reach$-states are absorbing. 
The maximal winning region is productive towards $\lifted{\reach}$ (and thus deadlock-free)  by definition.
Intuitively, while a deadlock-free region ensures that one never has to leave the region, 
any productive winning region ensures that from every belief support within this region there is a policy to stay in the winning region and 
that can almost-surely reach a $\reach$-state.
%
In particular, to find a winning policy (Challenge~1) or for the purpose of safe exploration (Challenge~2), it is sufficient to find a productive subset of the maximal winning region. We detail on this insight in the next subsection. 
\begin{center}
\fcolorbox{black}{orange!20!white!95!black}{
\parbox[l]{0.95\columnwidth}{
\textbf{Problem~3:} Given a POMDP $\pomdp$ and a specification $\varphi$, find a (large) productive winning region $\winning{\varphi}$.  	
}
}
\end{center}

To allow a compact representation
of winning regions, we exploit that 
for any belief support $\belsup' \subseteq \belsup$ it holds that $\post{\belsup'}{\act} \subseteq \post{\belsup}{\act}$ for all actions $\act\in\Act$, that is, the successors of $\belsup'$ are contained in the successors of $\belsup$.
\begin{lemma}
\label{lem:closedsubset}
	For winning belief support $\belsup$, $\belsup' \subseteq \belsup$ is winning.	
\end{lemma}

\subsection{Almost-Sure Reachability Shields in POMDPs}
We define a \emph{shield} -- towards the application of safe exploration -- that blocks actions which would lead an agent out of a winning region.
In particular, the shield imposes restrictions on policies to satisfy the reach-avoid specification. 
Technically, we adapt so-called \emph{permissive} policies~\cite{DBLP:journals/corr/DragerFK0U15,DBLP:conf/tacas/Junges0DTK16} for a belief-support MDP.
To force an agent to stay within a productive winning region $\winning{\varphi}$ for specification $\varphi$, we define a \emph{$\varphi$-shield} $\psched\colon \belsup \rightarrow 2^\Act$ such that  
for any winning $\belsup$ for $\varphi$ we have 
$\psched(\belsup) \subseteq \{ \act\in\Act \mid \post{\belsup}{\act} \subseteq \winning{\varphi} \}$, i.e., an action is part of the shield $\psched(\belsup)$ if it exclusively leads to belief support states within the winning region.

A shield $\psched$ restricts the set of actions an arbitrary policy may take\footnote{While memory policies based on the belief (support) are sufficient to ensure almost-sure reachability, the goal is to shield other policies that do not necessarily fall in this restricted class.}.
We call such restricted policies \emph{admissible}. 
Specifically, let $\belsup_\trace$ be the belief support after observing an observation sequence $\trace$. 
Then policy $\sched$ is $\psched$-admissible if $\support{\sched(\trace)} \subseteq \psched(\belsup_\trace)$ 
	for every observation-sequence $\trace$. Consequently, a policy is \emph{not} admissible if for some observation sequence $\trace$, the policy selects an action $\act\in\Act$ which is not allowed by the shield.

Some admissible policies may choose to stay in the winning region without progressing towards the $\reach$ states. Such a policy adheres to the avoid-part of the specification, but violates the reachability part.
 To enforce \emph{progress}, we adapt a notion of \emph{fairness}. 
A policy is fair if it takes every action infinitely often at any belief support state that appears infinitely often along a trace~\cite{BK08}. 
For example, a policy that randomizes (arbitrarily) over all actions is fair--we notice that most reinforcement learning policies are therefore fair.

\begin{theorem}
\label{thm:shielding}
For a $\varphi$-shield $\psched$ and a winning belief support $\belsup$, any \emph{fair} $\psched$-admissible policy satisfies $\varphi$ from $\belsup$.
\end{theorem}
\noindent We give a proof (sketch) in the appendix. The main idea is to show that the induced Markov chain of any admissable policy has only bottom SCCs that contain $\reach$-states.

\begin{remark}
If $\varphi$ is a safety specification (where $\probability_\belief^\sched(\avoid)=0$ suffices), we can rely on deadlock-free winning regions rather than productive winning regions and drop the fairness assumption.
\end{remark}

%
%
%
%

%% file: algorithm.tex
\section{Iterative SAT-Based Computation of Winning Regions}
\label{sec:computation_permissive}
We devise an approach for iteratively computing an increasing sequence of productive winning regions.
The approach delivers a compact symbolic encoding of winning regions: For a belief (or belief-support) state from a given winning region, we can efficiently decide whether the outcome of an action emanating from the state stays within the winning region. 

Key ingredient is the computation of so-called memoryless winning policies.  
We start this section by briefly recapping how to compute such policies directly on the POMDP, before we build an efficient incremental approach on top of this base method.
In particular, we first present a naive iterative algorithm based on the notion of \emph{shortcuts}, then describe how to implicitly add shortcuts within the encoding, and then finally combine the ideas to an efficient algorithm.

\subsection{One-shot approach to find small policies from a single belief}\label{sec:one_shot}
We aim to solve \textbf{Problem~1} and determine a winning policy.
The number of policies is exponential in the actions and the (exponentially many) belief support states. Searching among doubly exponentially many possibilities is intractable in general. 
However, Chatterjee et al.~\cite{DBLP:conf/aaai/ChatterjeeCD16} observe that often much simpler winning policies exist and provides a \emph{one-shot approach} to find them.
The essential idea is to search only for memoryless observation-based policies $\sched\colon \Obs \rightarrow \Distr(\Act)$ that are winning for the (initial) belief support $\belsup$.

\begin{figure}[t]
\centering
\subfigure[]{
\input{cheese1}
\label{fig:cheese:empty}
}
\qquad
\subfigure[]{
\input{cheese2}
\label{fig:cheese:memless}
}
\qquad
\subfigure[]{
\input{cheese3}
\label{fig:cheese:shortcut}
}
	\caption{Cheese-Maze example to explain memoryless policies and shortcuts }
	\label{fig:cheesepomdp}
\end{figure}

\begin{example}
\label{ex:cheesememless}
	Consider the small Cheese-POMDP~\cite{DBLP:conf/icml/LittmanCK95} in Fig.~\ref{fig:cheese:empty}. States are cells, actions are moving in the cardinal directions (if possible), and observations are the directions with adjacent cells, e.g., the boldface states $6,7,8$ share an observation. We set $\reach=\{10\}$ and $\avoid=\{9,11\}$.  From belief support $\belsup=\{ 6,8 \}$ there is no memoryless winning policy---In states $\{6,8\}$ we have to go north, which prevents us from going south in state $7$.
However, we can find a memoryless winning policy for $\{ 1,5 \}$, see Fig.~\ref{fig:cheese:memless}.
\end{example}
This problem is NP-complete, and it is thus natural to encode the problem as a satisfiability query in propositional logic.
We mildly adapt the original encoding of winning policies~\cite{DBLP:conf/aaai/ChatterjeeCD16}.
We introduce three sets of Boolean variables: $\encact{\obs}{\act}$, $\encreached{s}$ and $\encreach{s}{j}$.
If a policy takes action $\act \in \Act$ with positive probability upon observation $\obs \in \Obs$, then and only then, $\encact{\obs}{\act}$ is \texttt{true}.
If under this policy a state $s \in S$ is reached from some initial belief support $\belsup_\iota$ with positive probability, then and only then,  $\encreached{s}$ is \texttt{true}.
We define a maximal rank $\maxrank$ to ensure the productivity. 
For each state $s$ and rank $0 \leq j \leq \maxrank$, variable $\encreach{s}{j}$ indicates rank $j$ for $s$, that is, a path from $s$ leads to $s' \in \reach$ within $j$ steps.\footnote{Notice that a state $s$ can have multiple 'ranks' in this encoding. Its rank is the smallest $j$ such that $\encreach{s}{j}$ is \texttt{true}.}
A winning policy is then obtained by finding a satisfiable solution (via a SAT solver) to the conjunction $\encmemless{\pomdp}{\varphi}{\belsup_\iota}{k}$ of the constraints~\eqref{eq:init}--\eqref{eq:ranking}, where $\questionmark = S \setminus \big(\avoid \cup \reach \big)$. 

\begin{subequations}
  \begin{tabularx}{0.95\textwidth}{Xp{0.3cm}X}
  \begin{align}
    \bigwedge_{s \in \belsup_\iota} \encreached{s} \label{eq:init} 
     \end{align}
     & & 
  \begin{align}
    \bigwedge_{\obs \in \Obs}~\big(\bigvee_{\act \in \enact{\obs}} \encact{\obs}{\act} \big)\label{eq:reach} 
  \end{align}
  \end{tabularx}
\end{subequations}

	\noindent The initial belief support is clearly reachable~\eqref{eq:init}.
	 	The conjunction in \eqref{eq:reach} ensures that in every observation, at least one action is taken.  
	
	\begin{align}
	\bigwedge_{s \in \avoid} \neg \encreached{s} \quad\land \!\!\bigwedge_{\substack{s \in S\\\act \in \enact{s}}} \Bigl(
	 \encreached{s} \land \encact{\obsfun(s)}{\act} \rightarrow \bigwedge_{s' \in \post{s}{\act}} \encreached{s'} \Bigr)\label{eq:transreach} 
	 	\end{align}
	 	The conjunction~\eqref{eq:transreach} ensures  that the states in $B$ are in the set of reached states, that no states in $\avoid$ is reached,  and that this set is transitively closed under reachability (for the policy described by $\encact{\obs}{\act}$).	
	\begin{align}
	& \bigwedge_{s \in \questionmark} \encreached{s} \rightarrow  \encreach{s}{\maxrank} \label{eq:property} \\
	& \bigwedge_{s \not\in \reach} \neg\encreach{s}{0} \quad \land \bigwedge_{\substack{s \in \questionmark\\1 \leq j \leq \maxrank}}\!  \encreach{s}{j} \leftrightarrow \Bigl(\!\!\!\!\bigvee_{\act \in \enact{s}}\!\!\! \bigl( \encact{\obsfun(s)}{\act} \land \bigl(\!\!\!\!\bigvee_{\substack{s' \in \post{s}{\act}}}\!\!\!\!\encreach{s'}{j-1} \bigr) \!\bigr) \Bigr) \label{eq:ranking}
\end{align}
Conjunction \eqref{eq:property} states that any state that is reached almost-surely reaches a state in $\reach$, i.e., that there is a path (of length at most) $k$ to the target. 
Conjunctions \eqref{eq:ranking} describe a ranking function that ensures the existence of this path. 
Only states in $\reach$ have rank zero, and a state with positive probability to reach a state with rank $j{-}1$ within a step has rank at most $j$.  
%

By \cite[Thm.~2]{DBLP:conf/aaai/ChatterjeeCD16}, it holds that the conjunction $\encmemless{\pomdp}{\varphi}{\belsup_\iota}{k}$ of the constraints~\eqref{eq:init}--\eqref{eq:ranking}
is satisfiable, if there is a memoryless observation-based policy such that  $\varphi$ is satisfied.
If $\maxrank=|S|$, then the reverse direction also holds.
If $\maxrank < |S|$, we may miss states with a higher rank. 
Large values for $\maxrank$ are practically intractable~\cite{DBLP:conf/aaai/ChatterjeeCD16}, as the encoding grows significantly with $\maxrank$. 
Pandey and Rintanen~\cite{DBLP:conf/aips/PandeyR18} propose extending SAT-solvers with a dedicated handling of ranking constraints.


In order to apply this to small-memory policies, one can unfold $\log(m)$ bits of memory of such a policy into an $m$ times larger  POMDP~\cite{DBLP:conf/aaai/ChatterjeeCD16,junges2018finite}, and then search for a memoryless policy in this larger POMDP. Chatterjee et al.~\cite{DBLP:conf/aaai/ChatterjeeCD16} include a slight variation to this unfolding, allowing smaller-than-memoryless policies by enforcing the same action over various observations.

\subsection{Iterative Shortcuts}
\label{sec:idea}
We exploit the one-shot approach to create a naive iterative algorithm that constructs a  productive winning region.
The iterative algorithm avoids the following restrictions of the one-shot approach.
(1) In order to increase the likelihood of finding winning policies, we do not restrict ourselves to 
small-memory policies, and (2) we do not have to fix a maximal rank $\maxrank$.
These modifications allow us to find more winning policies, without guessing hyper-parameters. 
As we do not need to fix the belief-state, those parts of the winning region that are easy to find for the solver are encountered first.

\paragraph{The one-shot approach on winning regions.}
To understand the naive iterative algorithm, it is helpful to consider the previous encoding in the light of \textbf{Problem 3}, i.e., finding productive winning regions.
Consider first the interpretation of the variables. 
Indeed, observe that we have found \emph{the same} winning policy for all states $s$ where $C_s$ is \texttt{true}. Consequentially,
any belief support $\belsup_\obs = \{ s \mid C_s~\texttt{true} \land \obsfun(s) = \obs \}$ is winning.
\begin{lemma}
	If $\sched$ is winning for $\belsup$ and $\belsup'$, then $\sched$ is also winning for $\belsup \cup \belsup'$. 
\end{lemma}
\noindent This lemma is somewhat dual to Lemma~\ref{lem:closedsubset}, but requires a fixed policy. 
The constraints \eqref{eq:transreach} and  ensure that a winning-region is deadlock-free.  
The constraints \eqref{eq:property} and \eqref{eq:ranking} ensure productivity of the winning region. 

\paragraph{Adding shortcuts explicitly.}
The key idea is that we iteratively add \emph{short-cuts} in the POMDP that represent known winning policies. 
We find a winning policy $\sched$ for some belief states in the first iteration, and then add a fresh action $\act_\sched$ to all (original) POMDP states:
This action leads -- with probability one -- to a $\reach$ state, if the state is in the wining belief-support under policy $\sched$.
Otherwise, the action leads to an $\avoid$ state. 
\begin{definition}
For POMDP $\pomdp = \langle \mdp, \Obs, \obsfun \rangle$ where 
	$\mdp = \langle S, \Act, \dinit, \prob \rangle$
	and a policy $\sched$ with associated winning region $W_\varphi^\sched$, and assuming w.l.o.g., $\top \in \reach$ and $\bot \in \avoid$, we define the shortcutted POMDP $\pomdp\{\sched\} = \langle \mdp', \Obs, \obsfun \rangle$ where 
	$\mdp = \langle S, \Act', \dinit, \prob' \rangle$ with $\Act' = \Act \cup \{ \alpha_\sched \}$ and $\prob'(s,\act) = \prob(s,\act)$ for all $s \in S$ and $\act \in \Act$, and $\prob'(s,\act_\sched) = \{ \top \mapsto \iverson{\{ s \} \in W_\varphi^\sched}, \bot \mapsto \iverson{\{ s \} \not\in W_\varphi^\sched}\}$. 
\label{def:addshortcuts}
\end{definition}
\begin{lemma}
For a POMDP $\pomdp$ and policy $\sched$, the (maximal) winning regions for $\pomdp\{\sched\}$ and $\pomdp$ coincide.
\end{lemma}
\noindent First, adding more actions will not change a winning belief-support to be not winning.
Furthermore, by construction, taking the novel action will only lead to a winning belief-support whenever following $\sched$ from that point onwards would be a winning policy. 
The \emph{key} benefit is that adding shortcuts may extend the set of belief-support states that win via a memoryless policy.  This observation also gives rise to the following extension to the one-shot approach.
\begin{example}	
We continue with Example~\ref{ex:cheesememless}.
  If we add shortcuts, we can now find a memoryless winning policy for $\belsup=\{ 6,8 \}$, depicted in Fig.~\ref{fig:cheese:shortcut}.
\end{example}

\paragraph{Iterative shortcuts to extend a winning region.}
The idea is now to run the one-shot approach, extract the winning region, add the shortcuts to the POMDP, and rerun the one-shot approach. 
To make the one-shot approach applicable in this setting, it only needs one change: Rather than fixing an initial belief-support, we ask for an arbitrary new belief-support to be added to the states that we have previously covered. 
We use a data structure $\winningdata$ such that $\winningdata(\obs)$ encodes all winning belief supports with observation $\obs$.
Internally, the data structure stores maximal winning belief supports (w.r.t. set inclusion, see also Lemma~\ref{lem:closedsubset}) as bit-vectors.
By construction, for every $\belsup \in \winningdata(\obs)$, a winning region exists, i.e., conceptually, there is a shortcut-action leading to~$\reach$.

We extend the encoding (in partial preparation of the next subsection) and add a variable 
 $\encupdated{\obs} \in \belsup$ that is \texttt{true} if the policy is winning in a belief support that is not yet in $\winningdata(z)$.
We replace \eqref{eq:init} with:
\begin{align}
	& \biglor_{\obs \in \Obs} \encupdated{\obs}
\quad\land
\bigland_{\substack{\obs \in \Obs\\\winningdata(\obs) = \emptyset}}
\Bigl( \encupdated{\obs} \leftrightarrow \biglor_{\substack{s \in S\\\obsfun(s) = \obs}} \encreached{s} \Bigr)  \quad \land 
 \bigland_{\substack{\obs \in \Obs\\\winningdata(\obs) \neq \emptyset}} \Bigl( \encupdated{\obs} \leftrightarrow \bigland_{X \in \winningdata(\obs)}\biglor_{\substack{s \in S \setminus X\\\obsfun(s) = \obs}} \encreached{s} \Bigr)\label{eq:ext:progress1}
\end{align} 
For an observation $\obs$ for which we have not found a winning belief support yet, finding a policy from any state $s$ with $\obsfun(s)$ updates the winning region. 
Otherwise, it means finding a winning policy for a belief support that is not subsumed by a previous one~\eqref{eq:ext:progress1}.  

\paragraph{Real-valued ranking.}
To avoid setting a maximal path length, we use unbounded (real) variables $\encrank{s}$ rather than Boolean variables for the ranking~\cite{DBLP:journals/tcs/WimmerJAKB14}.
This relaxation avoids the growth of the encoding and admits arbitrarily large ranks with a fixed-size encoding into difference logic.  
 This logic is an extension to propositional logic that can be checked using an SMT solver~\cite{DBLP:series/faia/BarrettSST09}.
\begin{align}
	 &\bigwedge_{\substack{s \in \questionmark}}  \encreached{s} \rightarrow \Bigl(\!\!\!\!\bigvee_{\act \in \enact{s}}\!\!\!\!\!\bigl( \encact{\obsfun(s)}{\act} \land \bigl( \!\!\!\!\!\bigvee_{\substack{s' \in \post{s}{\act}}}\!\!\!\!\! \encrank{s} > \encrank{s'} \bigr)  \bigr)  \Bigr) \label{eq:naive:ranking} \end{align}
	 We replace~\eqref{eq:property} and \eqref{eq:ranking}: A state must have a successor state with a lower rank -- as before, but with real-valued ranks~\eqref{eq:naive:ranking}.

\begin{algorithm}[t]
\caption{Naive construction of winning regions}
\algtext*{EndWhile}
\footnotesize{
\begin{algorithmic}
\State \textbf{Input}: POMDP $\pomdp$, reach-avoid specification $\varphi$ \State \textbf{Output}: Winning region encoded in $\winningdata$
\State $\winningdata(\obs) \gets \{ s \in \reach \mid \obsfun(s) = \obs \} \text{ for all } \obs \in \Obs$
\State $\Phi \gets \textsl{Encode}(\pomdp,\varphi,\winningdata)$\Comment{Create encoding \eqref{eq:reach},\eqref{eq:transreach},\eqref{eq:ext:progress1},\eqref{eq:naive:ranking}.}

	\While{$\exists \eta \text{ s.t.\ } \eta \models \Phi$}\Comment{Call an SMT solver}
		\State $\winningdata(z) \gets \winningdata(z) \cup \{ \belsup \mid s \in \belsup \text{ iff } \eta(\encreached{s})  \} \text{ for all } \obs \in \Obs$
		\State $\pomdp \gets \pomdp\{\sched_\eta\}$ \Comment{Extend POMDP with Def.~\ref{def:addshortcuts}}
		\State \Comment{ with $\sched_\eta$ policy encoded by $\eta$.}
		\State $\Phi \gets  \textsl{Encode}(\pomdp,\varphi,\winningdata)$
	\EndWhile{}
\end{algorithmic}	
}
\label{alg:naiveexplicit}
\end{algorithm}

\paragraph{Algorithm.}
Together, the algorithm is given in Alg.~\ref{alg:naiveexplicit}. We initialize the winning region based on the specification, then encode the POMDP using the (modified) one-shot encoding. As long as the SMT solver finds policies that are winning for a new belief-support, we add those belief supports to the winning region. 
 In each iteration, $\winningdata$ contains a winning region. 
Once we find no more policies that extend the winning region on the  extended POMDP, we terminate.

The algorithm always terminates because the set of winning regions is finite, but in general does not solve \textbf{Problem 2}. 
Formally, the maximal winning region is a greatest fixpoint~\cite{BK08} and we iterate from below, i.e., the fixpoint that we find will be the smallest fixpoint (of the operation that we implement).  
However, iterating from above requires to reason that none of the doubly-exponentially many policies is winning for a particular belief support state; whereas our approach profits from finding simple strategies early on. 
Unfolding of memory as discussed earlier also makes this algorithm complete, yet, suffers from the same blow-up. 
A main advantage is that the algorithm often avoids the need for unfolding when searching for a winning policy or large winning regions.

In the following, we address two weaknesses: First, the algorithm currently creates a new encoding in every iteration, yielding significant overhead. Second, the algorithm in many settings requires adding a bit of memory to realize behavior where in a particular observation, we \emph{first} want to execute an action $\alpha$ and \emph{then} follow a shortcut from the state (with the same observation) reached from there. We adapt the encoding to explicitly allow for these  (non-memoryless) policies.

\subsection{Incremental encoding of winning regions}
\label{sec:incremental}
In this section, instead of naively adjusting the POMDP, we realize the idea of adding shortcuts  directly on the encoding. This encoding is the essential step towards an efficacious approach for solving \textbf{Problem 3}.
We find winning states based on a previous solution, and instead of adding actions, we allow the solver to decide following individual policies from each observation. In Section~\ref{sec:algorithm}, we embed this encoding into an improved algorithm.

Our encoding represents an observation-based policy that can decide to take a shortcut, which means that it follows a previously computed winning policy from there (implicitly using Lemma~\ref{lemma:winning_support_composable_policies}).
In addition to $\encact{\obs}{\act}$, $\encreached{s}$  and $\encrank{s}$ from the previous encoding, we use the following variables:
The policy takes shortcuts in states $s$ where $\encctd{s}$ is \texttt{true}.
For each observation, we must take the same shortcut, referred to by a positive integer-valued index  $\encprevpol{\obs}$. More precisely, $\encprevpol{\obs}$ refers to a shortcut from a previously computed (fragment of a) winning region stored in $\winningdata(\obs)_{\encprevpol{\obs}}$.  
The policy may decide to \emph{switch}, that is, to follow a shortcut \emph{after} taking an action starting in a state with observation~$\obs$. If $\encswitch{\obs}$  is \texttt{true}, the policy takes some action from $\obs$-states and from the next state, we take a shortcut. The encoding thus implicitly represents policies that are not memoryless but rather allow for a particular type of memory.

\noindent The conjunction of~\eqref{eq:ext:progress1} and \eqref{eq:ext:atleastoneaction}--\eqref{eq:ext:ranking} yields the encoding $\enciterative{\pomdp}{\varphi}{\winningdata}$:
\begin{align}
	& \bigwedge_{\obs \in \Obs} \Big(\biglor_{\act \in \enact{\obs}} \encact{\obs}{\act} \Big) \quad\land\quad \bigwedge_{s \in \avoid} \neg \encreached{s}\land \neg \encctd{s} \label{eq:ext:atleastoneaction}\\
	& \bigwedge_{\substack{s \in S\\\act \in \enact{s}}} \Bigl(
	 \encreached{s} \land \encact{\obsfun(s)}{\act} \land \neg\encswitch{\obsfun(s)} \quad\rightarrow\quad \!\!\!\bigwedge_{s' \in \post{s}{\act}}\!\!\! \encreached{s'} \Bigr) \label{eq:ext:reachnoswitch} \\
	 & \bigwedge_{\substack{s \in S\\\act \in \enact{s}}} 
	\Bigl(
	 \encreached{s} \land \encact{\obsfun(s)}{\act} \land \encswitch{\obsfun(s)} \quad\rightarrow \quad \bigwedge_{s' \in \post{s}{\act}} \encctd{s'}\Bigr)	\label{eq:ext:reachafterswitch} \end{align}
Similar to \eqref{eq:reach},\eqref{eq:transreach},
we select at least one action and \avoid-states should not be reached \eqref{eq:ext:atleastoneaction}. States reached are closed under the transitive closure, however, only if we do not switch to taking a shortcut~\eqref{eq:ext:reachnoswitch}. 
Furthermore, we mark the states reached after switching \eqref{eq:ext:reachafterswitch} and  need to select a shortcut for these states.
	 \begin{align}
	& \bigwedge_{s\in S} \big(\encctd{s} ~\rightarrow~ \encprevpol{\obsfun(s)} > 0\big) \quad \land\quad  \bigland_{\obs \in \Obs} \encprevpol{\obs} \leq |\winningdata(\obs)|\label{eq:ext:continue}\\
	& \bigland_{\substack{\obs\in \Obs\\0 \leq i \leq |\winningdata(z)|}}	 \bigland_{\substack{s \in S \setminus \winningdata(z)_i\\\obsfun(s) = \obs}} \encctd{s} ~\rightarrow~ \encprevpol{\obs} \neq i \label{eq:ext:continuewinning}\end{align}
If we reach a state $s$ after switching, then we must pick a shortcut. We can only pick an index that reflects a found winning region~\eqref{eq:ext:continue}.
If we pick this shortcut reflecting a winning region (fragment) for observation $\obs$, then we are winning from the states in $\winningdata(\obs)_i$, but not from any other state $s$ with that observation. 
Thus, for $s \not\in \winningdata(\obs)_i$, if we are going to follow any shortcut (that is, $\encctd{s}$ holds), we should not pick this particular shortcut encoded by $\encprevpol{\obs}$ (because it will lead to an $\avoid$-state). In terms of the policy: Taking this previously computed policy from state $s$ is not (known to) lead us to a $\reach$-state
\eqref{eq:ext:continuewinning}. 	Finally, we update the ranking to account for shortcuts.
\begin{align}
	 &\bigwedge_{\substack{s \in \questionmark}}  \encreached{s} \rightarrow \Bigl(\!\!\!\!\bigvee_{\act \in \enact{s}}\!\!\!\!\!\bigl( \encact{\obsfun(s)}{\act} \land \bigl( \!\!\!\!\!\bigvee_{\substack{s' \in \post{s}{\act}}}\!\!\!\!\! \encrank{s} > \encrank{s'} \bigr)  \bigr) \lor \encswitch{\obs(s)} \Bigr) \label{eq:ext:ranking} \end{align}
	 We make a slight adaption to \eqref{eq:naive:ranking}:  Either we have a successor state with a lower rank (as before) or we follow a shortcut---which either leads to the target or to a violation of the specification~\eqref{eq:ext:ranking}. 
	 
	 The correctness of the encoding is formalized below.
	 \begin{lemma}
If $\eta \models \enciterative{\pomdp}{\varphi}{\winningdata}$, then for every observation $\obs$, the belief support $\belsup_\obs = \{ s \mid \eta(\encreached{s}) = \texttt{true}, \obsfun(s) = \obs \}$ is winning.	
\end{lemma}
\noindent Algorithm~\ref{alg:naive} is a straightforward adaption of Algorithm~\ref{alg:naiveexplicit} that avoids adding shortcuts explicitly (and uses the updated encoding). As before, the algorithm terminates and solves \textbf{Problem 3}.
We conclude:
\begin{algorithm}[t]
\caption{Naive construction of winning regions with incremental encoding}
\algtext*{EndWhile}
\footnotesize{
\begin{algorithmic}
\State \textbf{Input}: POMDP $\pomdp$, reach-avoid specification $\varphi$
 \State \textbf{Output}: Winning region encoded in $\winningdata$
\State $\winningdata(\obs) \gets \{ s \in \reach \mid \obsfun(s) = \obs \} \text{ for all } \obs \in \Obs$
\State $\Phi \gets \textsl{Encode}(\pomdp,\varphi,\winningdata)$\Comment{Create encoding \eqref{eq:ext:progress1},\eqref{eq:ext:atleastoneaction}--\eqref{eq:ext:ranking}.}

	\While{$\exists \eta \text{ s.t.\ } \eta \models \Phi$}\Comment{Call an SMT solver}
		\State $\winningdata(z) \gets \winningdata(z) \cup \{ \belsup \mid s \in \belsup \text{ iff } \eta(\encreached{s})  \} \text{ for all } \obs \in \Obs$
		\State $\Phi \gets  \textsl{Encode}(\pomdp,\varphi,\winningdata)$
	\EndWhile{}
\end{algorithmic}	
}
\label{alg:naive}
\end{algorithm}
%
\begin{theorem}
 In any iteration, Algorithm~\ref{alg:naive} computes a  productive winning region.
\end{theorem}

\subsection{An Incremental Algorithm}
\label{sec:algorithm}
We adapt the algorithm sketched above to exploit the incrementality of modern SMT solvers. Furthermore, we aim to reduce the invocations of the solver by finding some trivial extensions to the winning region via a graph-based algorithm. 

\paragraph{Graph-based preprocessing.}
To reduce the number of SMT invocations, we employ polynomial-time graph-based heuristics.
The first step is to use (fully observable) MDP model checking on the POMDP as follows: find all states that under each (not necessarily observation-based) policy reach an $\avoid$-state with positive probability, and make them absorbing. 
Then, we find all states that under \emph{each} policy reach a $\reach$-state almost-surely.
Then, we iteratively search for \emph{winning observations} and use them to extend the $\reach$-states.
An observation $\obs$ is winning, if the belief-support $\{ s \mid \obsfun(s) = \obs\}$ is winning.
We start with a previously determined winning region $\winning{}$. 
We iteratively update $\winning{}$ by adding states $\belsup_\obs = \{ s \mid \obsfun(s) = \obs \}$ for some observation $\obs$, if there is an action $\act$ such that from every $s \in \belsup_\obs$, it holds $\post{s}{\act} \subseteq \winning{}$.
The iterative updates are interleaved with MDP model checking on the POMDP as described above until we find a fixpoint.

\paragraph{Optimized algorithm.}

\begin{algorithm}[t]
\caption{Incremental construction of winning regions}
\algtext*{EndWhile}
\footnotesize{
\begin{algorithmic}
\State \textbf{Input}: POMDP $\pomdp$, reach-avoid specification $\varphi$ \State \textbf{Output}: Winning region encoded in $\winningdata$
\State $\winningdata(\obs) \gets \{ s \in \reach \mid \obsfun(s) = \obs \} \text{ for all } \obs \in \Obs$
\State $\winningdata \gets \textsl{GraphPreprocessing}(\winningdata)$ 
\State $\Phi_\text{fix} \gets \textsl{Encode}_\text{fix}(\pomdp,\varphi,\winningdata)$\Comment{Create encoding \eqref{eq:ext:atleastoneaction}--\eqref{eq:ext:ranking}}
\State $\Phi_\text{inc} \gets \textsl{Encode}_\text{inc}(\pomdp,\varphi,\winningdata)$\Comment{Encode \eqref{eq:ext:progress1}}
\While{$\exists \eta \text{ s.t.\ } \eta \models \Phi$}\Comment{Call an SMT solver, fix $\eta$}

\Do \Comment{Extend policy}
\State $\Phi_\eta \gets \bigland \{ \encact{\obs}{\act} \mid \eta(\encupdated{\obs}) \land \eta(\encact{\obs}{\act}) \}$ \Comment{Part. fix policy}
\doWhile{$\exists \eta \text{ s.t.\ } \eta \models \Phi_\text{fix} \land \Phi_\text{var} \land \Phi_\eta$}\Comment{Call SMT, fix $\eta$}
	
		\State $\winningdata(z) \gets \winningdata(z) \cup \{ B \mid s \in B \text{ iff } \eta(\encreached{s})  \} \text{ for all } \obs \in \Obs$
		\State $\winningdata \gets \textsl{GraphPreprocessing}(\winningdata)$ 
		\State $\Phi_\text{fix} \gets  \Phi_\text{fix} \land \textsl{Encode}_{\eqref{eq:ext:continue}\eqref{eq:ext:continuewinning}}(\pomdp,\varphi,\winningdata)$\Comment{Update: \eqref{eq:ext:continue},\eqref{eq:ext:continuewinning} }
		\State $\Phi_\text{inc} \gets \textsl{Encode}_\text{inc}(\pomdp,\varphi,\winningdata)$\Comment{Encode \eqref{eq:ext:progress1}}
	\EndWhile{}
\end{algorithmic}	
}
\label{alg:optimised}
\end{algorithm}

We improve Algorithm~\ref{alg:naive} along four dimensions to obtain Algorithm~\ref{alg:optimised}.
First, we employ fewer updates of the winning region: We aim to extend the policy as much as possible, i.e., we want the SMT-solver to find more states with the same observation that are winning under the same policy.  
Therefore, we fix the variables for action choices that yield a new winning policy, and let the SMT solver search whether we can extend the corresponding winning region by finding more states and actions that are compatible with the partial policy.
Second, we observe that between (outer) iterations, large parts of the encoding stay intact, and use an incremental approach in which we first push all the constraints from the POMDP onto the stack, then all the constraints from the winning region, and finally a constraint that asks for progress. 
After we found a new policy, we pop the last constraint from the stack, add new constraints regarding the winning region (notice that the old constraints remain intact), and push new constraints that ask for extending the winning region to the stack. 
We refresh the encoding periodically to avoid unnecessary cluttering.
%
Third, further constraints (1) make the usage of shortcuts more flexible---we allow taking shortcuts either immediately or after the next action, and (2) enable an even more incremental encoding with some minor technical reformulations.
Fourth, we add the graph-preprocessing discussed above during the outer iteration.

%% file: cheese1.tex
\begin{tikzpicture}[scale=0.6]
\draw[thin] (0,3) rectangle node {\color{gray}1} (1,2);
\draw (1,3) rectangle node {\color{gray}2} (2,2);
\draw (2,3) rectangle node {\color{gray}3} (3,2);
\draw (3,3) rectangle node {\color{gray}4} (4,2);
\draw (4,3) rectangle node {\color{gray}5} (5,2);
\draw (0,2) rectangle node {\color{gray}\textbf{6}} (1,1);
\draw (0,1) rectangle node {\color{gray}9} (1,0);
\draw (2,2) rectangle node {\color{gray}\textbf{7}} (3,1);
\draw (2,1) rectangle node {\color{gray}10} (3,0);
\draw (4,2) rectangle node {\color{gray}\textbf{8}} (5,1);
\draw (4,1) rectangle node {\color{gray}11} (5,0);
\draw (0.5, 0.5) node[cross=6pt, draw=red] {};
\draw [draw=green, very thick](2.5, 0.5) circle (10pt) {};
\draw (4.5, 0.5) node[cross=6pt, draw=red] {};

\draw[line width=0.5mm,cheesebound] (0,0) edge (1,0);
\draw[line width=0.5mm,cheesebound] (0,0) edge (0,3);
\draw[line width=0.5mm,cheesebound] (5,3) edge (0,3);
\draw[line width=0.5mm,cheesebound] (1,2) edge (2,2);
\draw[line width=0.5mm,cheesebound] (1,2) edge (2,2);
\draw[line width=0.5mm,cheesebound] (3,2) edge (4,2);

\draw[line width=0.5mm,cheesebound] (2,0) edge (3,0);
\draw[line width=0.5mm,cheesebound] (4,0) edge (5,0);

\draw[line width=0.5mm,cheesebound] (5,0) edge (5,3);

\draw[line width=0.5mm,cheesebound] (4,0) edge (4,2);
\draw[line width=0.5mm,cheesebound] (3,0) edge (3,2);
\draw[line width=0.5mm,cheesebound] (2,0) edge (2,2);
\draw[line width=0.5mm,cheesebound] (1,0) edge (1,2);

\end{tikzpicture}

%% file: cheese2.tex
\begin{tikzpicture}[scale=0.6]
\draw[thin] (0,3) rectangle node {\color{gray}1} (1,2);
\draw (1,3) rectangle node {\color{gray}2} (2,2);
\draw (2,3) rectangle node {\color{gray}3} (3,2);
\draw (3,3) rectangle node {\color{gray}4} (4,2);
\draw (4,3) rectangle node {\color{gray}5} (5,2);
\draw (0,2) rectangle node {\color{gray}\textbf{6}} (1,1);
\draw (0,1) rectangle node {\color{gray}9} (1,0);
\draw (2,2) rectangle node {\color{gray}\textbf{7}} (3,1);
\draw (2,1) rectangle node {\color{gray}10} (3,0);
\draw (4,2) rectangle node {\color{gray}\textbf{8}} (5,1);
\draw (4,1) rectangle node {\color{gray}11} (5,0);
\draw (0.5, 0.5) node[cross=6pt, draw=red] {};
\draw [draw=green, very thick](2.5, 0.5) circle (10pt) {};
\draw (4.5, 0.5) node[cross=6pt, draw=red] {};

\draw[line width=0.5mm,cheesebound] (0,0) edge (1,0);
\draw[line width=0.5mm,cheesebound] (0,0) edge (0,3);
\draw[line width=0.5mm,cheesebound] (5,3) edge (0,3);
\draw[line width=0.5mm,cheesebound] (1,2) edge (2,2);
\draw[line width=0.5mm,cheesebound] (1,2) edge (2,2);
\draw[line width=0.5mm,cheesebound] (3,2) edge (4,2);

\draw[line width=0.5mm,cheesebound] (2,0) edge (3,0);
\draw[line width=0.5mm,cheesebound] (4,0) edge (5,0);

\draw[line width=0.5mm,cheesebound] (5,0) edge (5,3);

\draw[line width=0.5mm,cheesebound] (4,0) edge (4,2);
\draw[line width=0.5mm,cheesebound] (3,0) edge (3,2);
\draw[line width=0.5mm,cheesebound] (2,0) edge (2,2);
\draw[line width=0.5mm,cheesebound] (1,0) edge (1,2);

\draw[->,sched,very thick] (0.7, 2.6) -- (1.3, 2.6);
\draw[->,sched,very thick] (1.3, 2.4) -- (0.7, 2.4);
\draw[->,sched,very thick] (1.7, 2.5) -- (2.3, 2.5);
\draw[->,sched,very thick] (3.2, 2.5) -- (2.6, 2.5);
\draw[->,sched,very thick] (3.7, 2.6) -- (4.3, 2.6);
\draw[->,sched,very thick] (4.3, 2.4) -- (3.7, 2.4);
\draw[->,sched,very thick] (2.5, 2.2) -- (2.5, 1.7);
\draw[->,sched,very thick] (2.5, 1.2) -- (2.5, 0.7);
\draw[->,sched,very thick] (4.5, 1.2) -- (4.5, 0.7);
\draw[->,sched,very thick] (0.5, 1.2) -- (0.5, 0.7);
\end{tikzpicture}

%% file: cheese3.tex
\begin{tikzpicture}[scale=0.6]
\draw[thin] (0,3) rectangle node {\color{gray}1} (1,2);
\draw (1,3) rectangle node {\color{gray}2} (2,2);
\draw (2,3) rectangle node {\color{gray}3} (3,2);
\draw (3,3) rectangle node {\color{gray}4} (4,2);
\draw (4,3) rectangle node {\color{gray}5} (5,2);
\draw (0,2) rectangle node {\color{gray}\textbf{6}} (1,1);
\draw (0,1) rectangle node {\color{gray}9} (1,0);
\draw (2,2) rectangle node {\color{gray}\textbf{7}} (3,1);
\draw (2,1) rectangle node {\color{gray}10} (3,0);
\draw (4,2) rectangle node {\color{gray}\textbf{8}} (5,1);
\draw (4,1) rectangle node {\color{gray}11} (5,0);
\draw (0.5, 0.5) node[cross=6pt, draw=red] {};
\draw [draw=green, very thick](2.5, 0.5) circle (10pt) {};
\draw (4.5, 0.5) node[cross=6pt, draw=red] {};

\draw[line width=0.5mm,cheesebound] (0,0) edge (1,0);
\draw[line width=0.5mm,cheesebound] (0,0) edge (0,3);
\draw[line width=0.5mm,cheesebound] (5,3) edge (0,3);
\draw[line width=0.5mm,cheesebound] (1,2) edge (2,2);
\draw[line width=0.5mm,cheesebound] (1,2) edge (2,2);
\draw[line width=0.5mm,cheesebound] (3,2) edge (4,2);

\draw[line width=0.5mm,cheesebound] (2,0) edge (3,0);
\draw[line width=0.5mm,cheesebound] (4,0) edge (5,0);

\draw[line width=0.5mm,cheesebound] (5,0) edge (5,3);

\draw[line width=0.5mm,cheesebound] (4,0) edge (4,2);
\draw[line width=0.5mm,cheesebound] (3,0) edge (3,2);
\draw[line width=0.5mm,cheesebound] (2,0) edge (2,2);
\draw[line width=0.5mm,cheesebound] (1,0) edge (1,2);

\draw[->,sched,very thick] (2.5, 1.8) -- (2.5, 2.3);
\draw[->,sched,very thick] (0.5, 1.8) -- (0.5, 2.3);
\draw[->,sched,very thick] (4.5, 1.8) -- (4.5, 2.3);
\draw[->,dashed,prsched,very thick] (0.6, 2.4) -- (2.4, 0.6);
\draw[->,dashed,prsched,very thick] (4.4, 2.4) -- (2.6, 0.6);

\end{tikzpicture}

%% file: bdd.tex
\section{Model Checking the Belief-Support MDP}
\label{sec:belsupmodelchecking}

In this section, we briefly describe how we encode a given POMDP into a belief-support MDP to employ symbolic, off-the-shelf probabilistic model checking.
In particular, we employ symbolic (decision-diagram, DD) representations of the belief-support MDP as we expect this MDP to be huge. 
Constructing that DD representation effectively is not entirely trivial. 
Instead, we advocate constructing a (modular) symbolic description of the belief support MDP. 
Concretely, we automatically generate a model description in the MDP modeling language JANI~\cite{DBLP:conf/tacas/BuddeDHHJT17},\footnote{The description here works on a network of synchronized state machines as is also common in the \prism language} and then apply off-the-shelf model checking on the JANI description. 

Conceptually, we create a belief-support MDP with auxiliary states to allow for a concise encoding.\footnote{The usage of message passing or \emph{indexed assignments} in JANI would circumvent the need for intermediate states, but is to the best of our knowledge not supported by decision-diagram based model checkers.}  We use this auxiliary state $\hat{\belsup}$ to describe for any transition
 the conditioning on the observation. 
Concretely, a single transition $\prob(\belsup,\act,\belsup')$ in the belief-support MDP is reflected by two transitions $\prob(\belsup,\act,\hat{\belsup})$ and $\prob(\hat{\belsup},\act_\bot,\belsup')$ in our encoding, where $\act_\bot$ is a unique dummy action.
We encode states using triples $\langle\texttt{belsup},\texttt{newobs},\texttt{lact}\rangle$. 
$\texttt{belsup}$ is a bit vector with entries for every state $s$ that we use to encode the belief support. Variables $\texttt{newobs}$ and $\texttt{lact}$ store an observation and an action and are relevant only  for the auxiliary states.
Technically, we now encode the first transition from $\belsup$ with 
 the nondeterministic action $\act$ to $\hat{\belsup}$. 
$\prob(\belsup, \act)$ then yields (with arbitrary positive) probability a new observation that will reflect the observation $\obsfun(\belsup')$.  
We store $\act$ and $\obsfun(\belsup')$ in $\texttt{lact}$  and $\texttt{newobs}$, respectively. 
The second step is a single deterministic (dummy) action updating  $\texttt{belsup}$ while taking into account $\texttt{newobs}$. The step also resets $\texttt{lact}$ and $\texttt{newobs}$.  

The encoding of the transitions as follows: 
For the first step, we create nondeterministic choices for each action $\act$ and observation $\obs$. 
We guard these choices with $\obs$ meaning that the edge is only applicable to states having observation $\obs$, i.e., the guard is $\biglor_{\substack{s \in S,\obsfun(s) = \obs}} \texttt{belsup}(s)$. 
With these guarded edges, we define the destinations: With an arbitrary\footnote{We leave this a parametric probability in model building to reduce the number of different probabilities, as this is beneficial for the size of the decision diagram that \storm{} constructs -- it will only have leafs $0$, $p$, $1$.
Technically, such MDPs are not necessarily well-defined but we can employ model checking on the graph structure.} probability $p$, we go to an observation $\obs_1$ \emph{if} there is at least one state in $s \in \texttt{belsup}$ which has a successor state $s' \in \post{s}{\act}$ with $\obsfun(s') = \obs_1$.

The following pseudocode reflects the first step in the transition encoding.
The syntax is as follows: \textbf{take} an action \textbf{if} a Boolean guard is satisfied, then updates are executed with probability \textbf{prob}. 
An example for a guard is an observation $\obs$.
{\small
\[ \text{\textbf{take} $\act$ \textbf{if} $\obs$ \textbf{then} } 
\begin{cases} \textbf{prob } \big(\biglor_{\substack{s\in S \\ \prob(s,\act,\obs_1)>0}}  \texttt{belsup}(s) \text { ? }p\text{ : }0 \big) \colon & \begin{array}{@{}c@{}} \texttt{newobs} \gets \obs_1\\ \texttt{lact} \gets \act \end{array} 
\\
\hdots & \begin{array}{@{}c@{}}\hdots\end{array}  \\
\textbf{prob } \big(\biglor_{\substack{s\in S \\ \prob(s,\act,\obs_n)>0}}  \texttt{belsup}(s) \text { ? }p\text{ : }0\big) \colon & \begin{array}{@{}c@{}} \texttt{newobs} \gets \obs_n \\ \texttt{lact} \gets \act \end{array}  
\end{cases} 
\]}
The second step synchronously updates each state $s'$ in the POMDP independently: The entry $\texttt{belsup}(s')$ is set to \texttt{true} if $\obsfun(s) = \texttt{newobs}$ and if there is a state $s$ currently \texttt{true} in (the old) $\texttt{belsup}$ with $s' \in \post{s}{\texttt{lact}}$. The step thus can be captured by the following pseudocode for each $s'$: {\small\[ \text{\textbf{take} $\act_\bot$ \textbf{if} \texttt{true} \textbf{then} \textbf{prob} } 1: \texttt{belsup}(s') \gets \big( \biglor_{s} \prob(s,\texttt{lact},s')> 0 \big) \land \obsfun(s')\]} 
Finally, whenever the dummy action $\act_\bot$ is executed, we also reset the variables $\texttt{newobs}$ and $\texttt{lact}$.
The resulting encoding thus has transitions in the order of $|S| + |\Obs|^2\cdot|\max_{\obs \in \Obs} \enact{\obs}|$. 

%% file: experiments.tex
\section{Empirical Evaluation}
\label{sec:experiments}
We investigate the applicability of our incremental approach (Alg.~\ref{alg:optimised}) to \textbf{Challenge~1} and \textbf{Challenge~2}, and compare with our adaption and implementation of the one-shot approach~\cite{DBLP:conf/aaai/ChatterjeeCD16}, see Sect.~\ref{sec:one_shot}. 
We also employ the MDP model-checking approach from Sect.~\ref{sec:belsupmodelchecking}.

\paragraph{Setting.}
We implemented the one-shot algorithm, our incremental algorithm, and the generation of the JANI description of the belief support MDP on top of the model checker \storm{}~\cite{DBLP:conf/cav/DehnertJK017} and the  SMT solver \zthree{}~\cite{dMB08}. 
To compare with the one-shot algorithm for \textbf{Problem~1}, that is, for finding a policy from the \highlight{initial} state, we add a variant of Alg.~\ref{alg:optimised}.
Intuitively, any outer iteration starts with an SMT-check to see whether we find a policy covering the initial states. 
We realize the latter by fixing (temporarily) the $\encreached{s}$-variables. 
In the first iteration, this configuration and its resulting policy closely resemble the one-shot approach.
For the MDP model-checking approach, we use \storm{} (from the C++ API) with the \texttt{dd} engine and default  settings.   


For the experiments, we use a MacBook Pro MV962LL/A, a single core, no randomization, and use a 6GB  memory limit.
The time-out (TO) is 15 minutes.

%

\paragraph{Baseline.}
We compare with the one-shot algorithm including the graph-based preprocessing to identify more winning observations. 
We use two setups:
(1)~We (manually, a-priori) search for \highlight{optimal hyper-parameters} for each instance. We search for the smallest amount of memory possible, and for the smallest maximal rank $k$ (being a multiplicative of five) that yields a result. 
Guessing parameters as an ``oracle'' is time-consuming and unrealistic. 
We investigate (2)~the performance of the one-shot algorithm by \highlight{fixing the hyper-parameters} to two memory-states and $k=30$. 
These parameters provide results for most benchmarks. 
 
\paragraph{Benchmarks.} 
Our benchmarks involve agents operating in $N{\times}N$ grids, inspired by, \eg,~\cite{DBLP:conf/hybrid/SvorenovaCLECCB15,DBLP:conf/aaai/ChatterjeeCD16,Smith04heuristicsearch,DBLP:conf/icml/Dietterich98,DBLP:journals/corr/BrockmanCPSSTZ16}.
See Fig.~\ref{fig:screenshots} for video stills of simulating the following benchmarks.
\emph{Rocks} is a variant of \emph{rock sample}.
The grid contains two rocks which are either valuable or dangerous to collect. 
To find out with certainty, the rock has to be sampled from an adjacent field.
The goal is to collect a valuable rock, bring it to the drop-off zone, and not collect dangerous rocks.
\emph{Refuel} concerns a rover that shall travel from one corner to the other, while avoiding an obstacle on the diagonal.
Every movement costs energy and the rover may recharge at recharging stations to its full battery capacity $E$. 
It receives noisy information about its position and battery level.
\emph{Evade} is a scenario where a robot needs to reach a destination and evade a faster agent.
The robot has a limited range of vision ($R$), but may scan the whole grid instead of moving.
A certain safe area is only accessible by the robot.
\emph{Intercept} is inverse to \emph{Evade} in the sense that the robot aims to 
meet an agent before it leaves the grid via one of two available exits.
On top of the view radius, the agent observes a corridor in the center of the grid.
\emph{Avoid} is a related scenario where a robot shall keep distance to patrolling agents that move with uncertain speed, yielding partial information about their position
The robot may exploit their predefined routes.
\emph{Obstacle} contains static obstacles where the robot needs to reach the exit. 
Its initial state and movement are uncertain, and it only observes whether the current position is a trap or exit.

\input{table}

\paragraph{Results for Challenge 1.}
Tab.~\ref{tab:results} details the numerical benchmark results.
For each benchmark instance (columns), we report the name and relevant characteristics: the number of states ($|S|$), the number of transitions (\#Tr, the edges in the graph described by the POMDP), the number of observations ($|\Obs|$), and the number of belief support states ($|\belsup|$).
For the incremental method, we provide the run time (Time, in seconds), the number of outer iterations (\#Iter.) in Alg.~\ref{alg:optimised}, and the number of invocations of the SMT solver (\#solve), and the approximate size of the winning region ($|\winning{}|$). 
We then report these numbers when searching for a policy that wins from the initial state.
For the one-shot method, we provide the time for the optimal parameters (on the next line)--TOs reflect settings in which we did not find any suitable parameters, and the time for the preset parameters (2,30), or N/A if no policy can be found with these parameters.
Finally, for (belief-support) MDP model checking, we give only the run times.

The incremental algorithm finds winning policies for the \highlight{initial} state \emph{without guessing parameters} and is often \emph{faster} versus the one-shot approach with an oracle providing \highlight{optimal} parameters, and significantly faster than the one-shot approach with reasonably \highlight{fix}ed parameters. 
In detail, \emph{Rocks} shows that we can handle large numbers of iterations, solver invocations, and winning regions. 
The incremental approach scales to larger models, see e.g., \emph{Avoid}. 
\emph{Refuel} shows a large sensitivity of the one-shot method on the lookahead (going from 15 to 30 increases the runtime), while \emph{Evade} shows sensitivity to memory (from 1 to 2). 
In contrast, the incremental approach does not rely on user-input, yet delivers comparable performance on \emph{Refuel} or \emph{Avoid}. 
It suffers slightly on \emph{Evade}, where the one-shot approach has reduced overhead.
We furthermore conclude that off-the-shelf MDP model checking is not a fast alternative.
Its advantage is the guarantee to find the maximal winning region, however, for our benchmarks, maximal winning regions (empirically) coincide with the results from the incremental \highlight{fixpoint} approach. 

\paragraph{Results for Challenge 2.}
Winning regions obtained from running incrementally to a \highlight{fixpoint} are significantly larger than from running them until an \highlight{initial} winning policy is found (cf.\ the table), but requires some extra computational effort. 

If we let a \emph{shielded agent} move randomly through the grid-worlds, the  larger winning regions indeed translate to more permissiveness, that is, freedom to move for the agent (cf.\ the videos, Figure~\ref{fig:screenshots}). 
The shield is correct by construction, thus all runs indeed never reach an avoid state, and eventually reach the target (albeit after many steps). The latter is not true for the unshielded agents.

%
%

%% file: table.tex
\begin{table*}[t]
\centering
\caption{Numerical results towards solving \textbf{Problem~1} and \textbf{Problem~3}.}
\setlength{\tabcolsep}{2pt} 
\renewcommand{\arraystretch}{1.1} 
\footnotesize{
\scalebox{0.8}{
\begin{tabular}{cc|l|rr|rr|rr|rr|rr|rr}
&&           & \multicolumn{2}{c|}{\emph{Rocks} ($N$)} & \multicolumn{2}{c|}{\emph{Refuel} ($N$,$E$)}   & \multicolumn{2}{c|}{\emph{Evade} ($N$,$R$)}  &  \multicolumn{2}{c|}{\emph{Avoid} ($N$,$R$)}   & \multicolumn{2}{c|}{\emph{Intercept} ($N$,$R$)} & \multicolumn{2}{c}{\emph{Obstacle} ($N$)}  \\
&&Inst.            	& 4		& 6 	& 6,8 	& 7,7& 6,2   & 7,2  & 6,3 	& 7,4 	& 7,1 	& 7,2 	& 6		& 8 	\\\hline
&&$|S|$       	   	& 331 	& 816 	& 270 	& 302 & 4232  & 8108  	& 5976 	& 13021 & 4705 	& 4705 	& 37 	& 65	 \\
&&\#Tr  		   	& 3484 	& 7292		& 1301	& 1545 & 28866 & 57570 & 14373 & 33949 & 18049 & 18049 & 224 	& 421 	\\
&&$|\Obs|$  		& 65 	& 74 	& 36 	& 35 & 2202  	& 4172  	& 3300  & 8584 	& 2002 	& 2598 	& 4 	& 4 	  \\
&&$|\belsup|$  		& 3.5\oom{5} 	&  7.7\oom{25} & 5.6\oom{14}& 7.4\oom{19}  & 1.1\oom{8} & 4.4\oom{11} &  1.1\oom{15} & 2.9\oom{17} & 6.4\oom{10} & 2.7\oom{9} &1.1\oom{9} & 2.9\oom{17}  \\\hline\hline
\multirow{8}{*}{\rotatebox{90}{\textbf{incremental}}}&\multirow{4}{*}{\rotatebox{90}{\highlight{fixpoint}}}& \runtime{Time} 
       	 			& \runtime{19} 	& \runtime{753} 	& \runtime{6} 	& \runtime{3}& \runtime{142} 	& \runtime{613} 	& \runtime{167} 	& \runtime{745} 	& \runtime{116} 	& \runtime{86} 	& \runtime{2} 	& \runtime{30} 	 \\
&&\#Iter. 		 	& 36 	& 284 	& 40 	& 30  & 4     	& 6 		& 3 		& 4 		& 8 	& 8 		& 68 	& 150 	\\
&&\#solve    			& 1702 	& 13650 & 1023 	& 528 & 681   & 1129 	& 629  	& 1027 	& 1171 	& 976  & 839 	& 4291   \\
&&$|\winning{}|$ 	& 3.5\oom{5} & 7.7\oom{25} & 1.2\oom{11} &  2.1\oom{8} & 1.0\oom{8} & 4.2\oom{11}  &1.1\oom{15} & 2.9\oom{17} & 9.2\oom{4} & 2.9\oom{4}& 4.1\oom{7}& 3.8\oom{14}  \\\cline{2-15}
&\multirow{4}{*}{\rotatebox{90}{\highlight{initial}}}& \runtime{Time} 
			       	& \runtime{17} 	& \runtime{226}  	& \runtime{2} 	& \runtime{2} & \runtime{49} 	& \runtime{576}  	& \runtime{10} 	& \runtime{40} 	& \runtime{11} 	& \runtime{2} 		& \runtime{$<$1	} & \runtime{$<$1} 	 \\
&&\#Iter. 		 	& 29  	& 65 	&  2	& 4 & 1		& 1		& 1 		& 1 		& 2 	& 1 		& 10 	& 12 \\
&&\#solve    		& 1215 	& 2652 	& 62 	& 80 & 1		& 1		&  1 	&  1 	& 81 	& 1		& 114 	& 229 	 \\
&&$|\winning{}|$ 				& 4.4\oom{4} & 1.8\oom{13} & 8.4\oom{6} & 3.7\oom{4} &5.0\oom{7} & 1.0\oom{11}&  3.7\oom{5} & 6.9\oom{10} & 6.2\oom{3}  & 2.1\oom{3}&4.1\oom{5} & 4.5\oom{9}  \\\hline\hline
\multirow{3}{*}{\rotatebox{90}{\textbf{1-shot\,}}}
&\multirow{2}{*}{\rotatebox{90}{\highlight{opt\,}}}
&\runtime{Time} 				& \runtime{120} 	& \runtime{TO} 	& \runtime{2} 	& \runtime{$<$1}& \runtime{12}		& \runtime{270} 	& \runtime{22} 	& \runtime{53}  	& \runtime{8} 	& \runtime{1}  	& \runtime{1}  	& \runtime{195} 	 \\
&&Mem,k 			& 2,10 	& ? 	& 2,15 	& 2,15 & 1,20 	& 1,30 	& 1,30 	& 1,25 	& 2,10 	& 1,10	& 6,10 	& 5,50	\\\cline{2-15}
&\rotatebox{90}{\highlight{fix\,}}& 
\runtime{Time} 				& \runtime{TO} 	& \runtime{TO} 	& \runtime{11}  & \runtime{37} & \runtime{TO} 	& \runtime{TO}  	& \runtime{TO}  	& \runtime{TO} 	& \runtime{28} 	& \runtime{18} 	& N/A 	& N/A 	\\\hline\hline
\multicolumn{2}{c|}{\textbf{MDP}} & \runtime{Time}&  \runtime{400} 	& \runtime{TO}	& \runtime{219}  & \runtime{MO} & \runtime{TO}  	& \runtime{TO}  	& \runtime{TO}  	& \runtime{TO}	& \runtime{TO} 	& \runtime{TO} 	& \runtime{6} 	& \runtime{MO}  \\
\end{tabular}}
}
\label{tab:results}
\end{table*}

%% file: conclusion.tex
\section{Conclusion}
We provided an incremental approach to find POMDP policies that satisfy almost-sure reachability specifications.
The superior scalability is demonstrated on a string of  benchmarks. 
Furthermore, this approach allows to shield agents in POMDPs and guarantees that any exploration of an environment satisfies the specification, without needlessly restricting the freedom of the agent.
We plan to investigate a tight interaction with state-of-the-art reinforcement learning and quantitative verification of POMDPs.


%% file: proof.tex
\section{Proof Sketch of Theorem~\ref{thm:shielding}}
\noindent \emph{Proof sketch.}
By counterexample. Assume that a policy $\sched$ exists that does not satisfy $\varphi$. 

First, observe that as the policy $\psched$-admissable, there is no path from $\belsup$ to an $\avoid$ state. 
Thus, to violate $\varphi$, it must with positive probability evade arriving at $\reach$. Let us consider the induced Markov chain of $\sched$. In any (induced) MC, bottom SCCs are reached with probability $1$. We show below that there are only  bottom SCCs that include some $\reach$ state, which would lead to a contradiction and thus a proof. 

It remains to show that there is no other bottom SCC.  Again, by counterexample, we assume that there is another bottom SCC including some belief $\belsup_1$. Due to the productiveness of the winning region, there is a fixed path $\pi$ that leads to $\reach$, but then $\reach$ must be in the same bottom SCC. Contradiction. 
\qed

%% file: main.bbl
\begin{thebibliography}{10}

\bibitem{DBLP:conf/cdc/AkametaluKFZGT14}
Anayo~K. Akametalu, Shahab Kaynama, Jaime~F. Fisac, Melanie~Nicole Zeilinger,
  Jeremy~H. Gillula, and Claire~J. Tomlin.
\newblock Reachability-based safe learning with {G}aussian processes.
\newblock In {\em {CDC}}, pages 1424--1431. {IEEE}, 2014.

\bibitem{shield_rl}
Mohammed Alshiekh, Roderick Bloem, R{\"{u}}diger Ehlers, Bettina
  K{\"{o}}nighofer, Scott Niekum, and Ufuk Topcu.
\newblock Safe reinforcement learning via shielding.
\newblock In {\em {AAAI}}. {AAAI} Press, 2018.

\bibitem{DBLP:journals/aamas/AmatoBZ10}
Christopher Amato, Daniel~S. Bernstein, and Shlomo Zilberstein.
\newblock Optimizing fixed-size stochastic controllers for {POMDP}s and
  decentralized {POMDP}s.
\newblock {\em Auton. Agents Multi Agent Syst.}, 21(3):293--320, 2010.

\bibitem{DBLP:journals/jacm/BaierGB12}
Christel Baier, Marcus Gr{\"{o}}{\ss}er, and Nathalie Bertrand.
\newblock Probabilistic {\(\omega\)}-automata.
\newblock {\em J. {ACM}}, 59(1):1:1--1:52, 2012.

\bibitem{BK08}
Christel Baier and Joost-Pieter Katoen.
\newblock {\em Principles of Model Checking}.
\newblock MIT Press, 2008.

\bibitem{DBLP:series/faia/BarrettSST09}
Clark~W. Barrett, Roberto Sebastiani, Sanjit~A. Seshia, and Cesare Tinelli.
\newblock Satisfiability modulo theories.
\newblock In {\em Handbook of Satisfiability}, volume 185 of {\em Frontiers in
  Artificial Intelligence and Applications}, pages 825--885. {IOS} Press, 2009.

\bibitem{barrett-smtbookch09}
Clark~W. Barrett and Cesare Tinelli.
\newblock Satisfiability modulo theories.
\newblock In {\em Handbook of Model Checking}, pages 305--343. Springer, 2018.

\bibitem{DBLP:conf/aips/BertoliCP06}
Piergiorgio Bertoli, Alessandro Cimatti, and Marco Pistore.
\newblock Towards strong cyclic planning under partial observability.
\newblock In {\em {ICAPS}}, pages 354--357. {AAAI}, 2006.

\bibitem{DBLP:series/faia/2009-185}
Armin Biere, Marijn Heule, Hans van Maaren, and Toby Walsh, editors.
\newblock {\em Handbook of Satisfiability}, volume 185 of {\em Frontiers in
  Artificial Intelligence and Applications}. {IOS} Press, 2009.

\bibitem{DBLP:journals/corr/abs-2006-16688}
Roderick Bloem, Peter~Gj{\o}l Jensen, Bettina K{\"{o}}nighofer, Kim~Guldstrand
  Larsen, Florian Lorber, and Alexander Palmisano.
\newblock It's time to play safe: Shield synthesis for timed systems.
\newblock {\em CoRR}, abs/2006.16688, 2020.

\bibitem{DBLP:conf/tacas/BloemKKW15}
Roderick Bloem, Bettina K{\"{o}}nighofer, Robert K{\"{o}}nighofer, and Chao
  Wang.
\newblock Shield synthesis: - runtime enforcement for reactive systems.
\newblock In {\em {TACAS}}, volume 9035 of {\em {LNCS}}, pages 533--548.
  {Springer}, 2015.

\bibitem{DBLP:conf/atva/BorkJKQ20}
Alexander Bork, Sebastian Junges, Joost{-}Pieter Katoen, and Tim Quatmann.
\newblock Verification of indefinite-horizon pomdps.
\newblock In {\em {ATVA}}, volume 12302 of {\em Lecture Notes in Computer
  Science}, pages 288--304. Springer, 2020.

\bibitem{DBLP:journals/corr/BrockmanCPSSTZ16}
Greg Brockman, Vicki Cheung, Ludwig Pettersson, Jonas Schneider, John Schulman,
  Jie Tang, and Wojciech Zaremba.
\newblock Open{AI} {G}ym.
\newblock {\em CoRR}, abs/1606.01540, 2016.

\bibitem{DBLP:conf/tacas/BuddeDHHJT17}
Carlos~E. Budde, Christian Dehnert, Ernst~Moritz Hahn, Arnd Hartmanns,
  Sebastian Junges, and Andrea Turrini.
\newblock {JANI:} quantitative model and tool interaction.
\newblock In {\em {TACAS} {(2)}}, volume 10206 of {\em Lecture Notes in
  Computer Science}, pages 151--168, 2017.

\bibitem{burns2007sampling}
Brendan Burns and Oliver Brock.
\newblock {Sampling-Based Motion Planning with Sensing Uncertainty}.
\newblock In {\em ICRA}, pages 3313--3318. IEEE, 2007.

\bibitem{DBLP:conf/ijcai/CarrJT20}
Steven Carr, Nils Jansen, and Ufuk Topcu.
\newblock Verifiable rnn-based policies for pomdps under temporal logic
  constraints.
\newblock In {\em {IJCAI}}, pages 4121--4127. ijcai.org, 2020.

\bibitem{DBLP:conf/ijcai/Carr0WS0T19}
Steven Carr, Nils Jansen, Ralf Wimmer, Alexandru~Constantin Serban, Bernd
  Becker, and Ufuk Topcu.
\newblock Counterexample-guided strategy improvement for {POMDP}s using
  recurrent neural networks.
\newblock In {\em {IJCAI}}, pages 5532--5539. ijcai.org, 2019.

\bibitem{DBLP:conf/aaai/ChatterjeeCD16}
Krishnendu Chatterjee, Martin Chmelik, and Jessica Davies.
\newblock A symbolic {SAT}-based algorithm for almost-sure reachability with
  small strategies in {POMDPs}.
\newblock In {\em {AAAI}}, pages 3225--3232. {AAAI} Press, 2016.

\bibitem{DBLP:conf/icra/ChatterjeeCGK15}
Krishnendu Chatterjee, Martin Chmelik, Raghav Gupta, and Ayush Kanodia.
\newblock Qualitative analysis of {POMDP}s with temporal logic specifications
  for robotics applications.
\newblock In {\em {ICRA}}, pages 325--330. {IEEE}, 2015.

\bibitem{DBLP:journals/ai/ChatterjeeCGK16}
Krishnendu Chatterjee, Martin Chmelik, Raghav Gupta, and Ayush Kanodia.
\newblock Optimal cost almost-sure reachability in {POMDP}s.
\newblock {\em Artif. Intell.}, 234:26--48, 2016.

\bibitem{DBLP:conf/mfcs/ChatterjeeDH10}
Krishnendu Chatterjee, Laurent Doyen, and Thomas~A. Henzinger.
\newblock Qualitative analysis of partially-observable markov decision
  processes.
\newblock In {\em {MFCS}}, volume 6281 of {\em {LNCS}}, pages 258--269.
  Springer, 2010.

\bibitem{dMB08}
Leonardo~Mendon{\c{c}}a de~Moura and Nikolaj Bj{\o}rner.
\newblock {Z3:} an efficient {SMT} solver.
\newblock In {\em {TACAS}}, volume 4963 of {\em {LNCS}}, pages 337--340.
  Springer, 2008.

\bibitem{DBLP:conf/cav/DehnertJK017}
Christian Dehnert, Sebastian Junges, Joost-Pieter Katoen, and Matthias Volk.
\newblock A storm is coming: {A} modern probabilistic model checker.
\newblock In {\em {CAV} {(2)}}, volume 10427 of {\em {LNCS}}, pages 592--600.
  Springer, 2017.

\bibitem{DBLP:conf/icml/Dietterich98}
Thomas~G. Dietterich.
\newblock The {MAXQ} method for hierarchical reinforcement learning.
\newblock In {\em {ICML}}, pages 118--126. Morgan Kaufmann, 1998.

\bibitem{draeger-et-al-tacas-2014}
Klaus Dr{\"{a}}ger, Vojtech Forejt, Marta~Z. Kwiatkowska, David Parker, and
  Mateusz Ujma.
\newblock Permissive controller synthesis for probabilistic systems.
\newblock volume 8413 of {\em {LNCS}}, pages 531--546. {Springer}, 2014.

\bibitem{DBLP:journals/corr/DragerFK0U15}
Klaus Dr{\"{a}}ger, Vojtech Forejt, Marta~Z. Kwiatkowska, David Parker, and
  Mateusz Ujma.
\newblock Permissive controller synthesis for probabilistic systems.
\newblock {\em Logical Methods in Computer Science}, 11(2), 2015.

\bibitem{DBLP:conf/aaai/FultonP18}
Nathan Fulton and Andr{\'{e}} Platzer.
\newblock Safe reinforcement learning via formal methods: Toward safe control
  through proof and learning.
\newblock In {\em {AAAI}}, pages 6485--6492. {AAAI} Press, 2018.

\bibitem{DBLP:journals/jmlr/GarciaF15}
Javier Garc{\'{\i}}a and Fernando Fern{\'{a}}ndez.
\newblock A comprehensive survey on safe reinforcement learning.
\newblock {\em J. Mach. Learn. Res.}, 16:1437--1480, 2015.

\bibitem{DBLP:conf/tacas/HahnPSSTW19}
Ernst~Moritz Hahn, Mateo Perez, Sven Schewe, Fabio Somenzi, Ashutosh Trivedi,
  and Dominik Wojtczak.
\newblock Omega-regular objectives in model-free reinforcement learning.
\newblock In {\em {TACAS} {(1)}}, volume 11427 of {\em {LNCS}}, pages 395--412.
  Springer, 2019.

\bibitem{DBLP:conf/atal/HasanbeigAK20}
Mohammadhosein Hasanbeig, Alessandro Abate, and Daniel Kroening.
\newblock Cautious reinforcement learning with logical constraints.
\newblock In {\em {AAMAS}}, pages 483--491. International Foundation for
  Autonomous Agents and Multiagent Systems, 2020.

\bibitem{hausknecht2015deep}
Matthew~J. Hausknecht and Peter Stone.
\newblock Deep recurrent {Q}-learning for partially observable {MDP}s.
\newblock In {\em {AAAI}}, pages 29--37. {AAAI} Press, 2015.

\bibitem{hauskrecht2000value}
Milos Hauskrecht.
\newblock Value-function approximations for partially observable {M}arkov
  decision processes.
\newblock {\em J. Artif. Intell. Res.}, 13:33--94, 2000.

\bibitem{DBLP:conf/ijcai/HorakBC18}
Karel Hor{\'{a}}k, Branislav Bosansk{\'{y}}, and Krishnendu Chatterjee.
\newblock Goal-{HSVI}: Heuristic search value iteration for goal {POMDP}s.
\newblock In {\em {IJCAI}}, pages 4764--4770. ijcai.org, 2018.

\bibitem{DBLP:conf/nips/JaakkolaSJ94}
Tommi~S. Jaakkola, Satinder~P. Singh, and Michael~I. Jordan.
\newblock Reinforcement learning algorithm for partially observable markov
  decision problems.
\newblock In {\em {NIPS}}, pages 345--352. {MIT} Press, 1994.

\bibitem{jansen-et-al-shield}
Nils Jansen, Bettina K{\"{o}}nighofer, Sebastian Junges, Alex Serban, and
  Roderick Bloem.
\newblock Safe reinforcement learning using probabilistic shields (invited
  paper).
\newblock In {\em {CONCUR}}, volume 171 of {\em LIPIcs}, pages 3:1--3:16.
  Schloss Dagstuhl - Leibniz-Zentrum f{\"{u}}r Informatik, 2020.

\bibitem{junges-et-al-tacas-2016}
Sebastian Junges, Nils Jansen, Christian Dehnert, Ufuk Topcu, and Joost-Pieter
  Katoen.
\newblock Safety-constrained reinforcement learning for {MDPs}.
\newblock In {\em {TACAS}}, volume 9636 of {\em {LNCS}}, pages 130--146.
  {Springer}, 2016.

\bibitem{DBLP:conf/tacas/Junges0DTK16}
Sebastian Junges, Nils Jansen, Christian Dehnert, Ufuk Topcu, and Joost-Pieter
  Katoen.
\newblock Safety-constrained reinforcement learning for {MDP}s.
\newblock In {\em {TACAS}}, volume 9636 of {\em {LNCS}}, pages 130--146.
  {Springer}, 2016.

\bibitem{junges2018finite}
Sebastian Junges, Nils Jansen, Ralf Wimmer, Tim Quatmann, Leonore Winterer,
  Joost-Pieter Katoen, and Bernd Becker.
\newblock Finite-state controllers of {POMDP}s using parameter synthesis.
\newblock In {\em {UAI}}, pages 519--529. {AUAI} Press, 2018.

\bibitem{kaelbling1998planning}
Leslie~Pack Kaelbling, Michael~L. Littman, and Anthony~R. Cassandra.
\newblock Planning and acting in partially observable stochastic domains.
\newblock {\em Artif. Intell.}, 101(1-2):99--134, 1998.

\bibitem{DBLP:conf/icml/LittmanCK95}
Michael~L. Littman, Anthony~R. Cassandra, and Leslie~Pack Kaelbling.
\newblock Learning policies for partially observable environments: Scaling up.
\newblock In {\em {ICML}}, pages 362--370. Morgan Kaufmann, 1995.

\bibitem{MadaniHC99}
Omid Madani, Steve Hanks, and Anne Condon.
\newblock On the undecidability of probabilistic planning and infinite-horizon
  partially observable {Markov} decision problems.
\newblock In {\em AAAI}, pages 541--548. {AAAI} Press, 1999.

\bibitem{meuleau1999solving}
Nicolas Meuleau, Kee-Eung Kim, Leslie~Pack Kaelbling, and Anthony~R Cassandra.
\newblock Solving {POMDPs} by searching the space of finite policies.
\newblock In {\em {UAI}}, pages 417--426. Morgan Kaufmann Publishers Inc.,
  1999.

\bibitem{DBLP:journals/tsmc/NamA10}
Wonhong Nam and Rajeev Alur.
\newblock Active learning of plans for safety and reachability goals with
  partial observability.
\newblock {\em {IEEE} Trans. Syst. Man Cybern. Part {B}}, 40(2):412--420, 2010.

\bibitem{DBLP:journals/rts/Norman0Z17}
Gethin Norman, David Parker, and Xueyi Zou.
\newblock Verification and control of partially observable probabilistic
  systems.
\newblock {\em Real-Time Systems}, 53(3):354--402, 2017.

\bibitem{DBLP:conf/aips/PandeyR18}
Binda Pandey and Jussi Rintanen.
\newblock Planning for partial observability by {SAT} and graph constraints.
\newblock In {\em {ICAPS}}, pages 190--198. {AAAI} Press, 2018.

\bibitem{DBLP:conf/mesas/PeckaS14}
Martin Pecka and Tom{\'{a}}s Svoboda.
\newblock Safe exploration techniques for reinforcement learning - an overview.
\newblock In {\em {MESAS}}, volume 8906 of {\em {LNCS}}, pages 357--375.
  Springer, 2014.

\bibitem{pineau2003point}
Joelle Pineau, Geoff Gordon, and Sebastian Thrun.
\newblock Point-based value iteration: {An} anytime algorithm for {POMDPs}.
\newblock In {\em IJCAI}, pages 1025--1032. Morgan Kaufmann, 2003.

\bibitem{Pnueli77}
Amir Pnueli.
\newblock The temporal logic of programs.
\newblock In {\em FOCS}, pages 46--57. {IEEE} Computer Society, 1977.

\bibitem{DBLP:conf/nips/PoupartB03}
Pascal Poupart and Craig Boutilier.
\newblock Bounded finite state controllers.
\newblock In {\em {NIPS}}, pages 823--830. {MIT} Press, 2003.

\bibitem{Put94}
Martin~L. Puterman.
\newblock {\em {{M}arkov} Decision Processes}.
\newblock John Wiley and Sons, 1994.

\bibitem{DBLP:journals/lmcs/RaskinCDH07}
Jean{-}Fran{\c{c}}ois Raskin, Krishnendu Chatterjee, Laurent Doyen, and
  Thomas~A. Henzinger.
\newblock Algorithms for omega-regular games with imperfect information.
\newblock {\em Log. Methods Comput. Sci.}, 3(3), 2007.

\bibitem{ShaniPK13}
Guy Shani, Joelle Pineau, and Robert Kaplow.
\newblock A survey of point-based {POMDP} solvers.
\newblock {\em Autonomous Agents and Multi-Agent Systems}, 27(1):1--51, 2013.

\bibitem{silver2010monte}
David Silver and Joel Veness.
\newblock Monte-carlo planning in large {POMDPs}.
\newblock In {\em NIPS}, pages 2164--2172, 2010.

\bibitem{Smith04heuristicsearch}
Trey Smith and Reid Simmons.
\newblock Heuristic search value iteration for {POMDP}s, 2004.

\bibitem{DBLP:conf/hybrid/SvorenovaCLECCB15}
Mar{\'{\i}}a Svorenov{\'{a}}, Martin Chmelik, Kevin Leahy, Hasan~Ferit Eniser,
  Krishnendu Chatterjee, Ivana Cern{\'{a}}, and Calin Belta.
\newblock Temporal logic motion planning using {POMDP}s with parity objectives:
  case study paper.
\newblock In {\em {HSCC}}, pages 233--238. {ACM}, 2015.

\bibitem{thrun2005probabilistic}
Sebastian Thrun, Wolfram Burgard, and Dieter Fox.
\newblock {\em Probabilistic Robotics}.
\newblock The MIT Press, 2005.

\bibitem{DBLP:conf/nips/TurchettaB019}
Matteo Turchetta, Felix Berkenkamp, and Andreas Krause.
\newblock Safe exploration for interactive machine learning.
\newblock In {\em NeurIPS}, pages 2887--2897, 2019.

\bibitem{DBLP:conf/aaai/WalravenS17}
Erwin Walraven and Matthijs T.~J. Spaan.
\newblock Accelerated vector pruning for optimal {POMDP} solvers.
\newblock In {\em {AAAI}}, pages 3672--3678. {AAAI} Press, 2017.

\bibitem{DBLP:conf/atal/WangCK18}
Yue Wang, Swarat Chaudhuri, and Lydia~E. Kavraki.
\newblock Bounded policy synthesis for {POMDP}s with safe-reachability
  objectives.
\newblock In {\em {AAMAS}}, pages 238--246. International Foundation for
  Autonomous Agents and Multiagent Systems, 2018.

\bibitem{wierstra2007solving}
Daan Wierstra, Alexander F{\"o}rster, Jan Peters, and J{\"u}rgen Schmidhuber.
\newblock Solving deep memory {POMDPs} with recurrent policy gradients.
\newblock In {\em ICANN}, pages 697--706. Springer, 2007.

\bibitem{DBLP:journals/tcs/WimmerJAKB14}
Ralf Wimmer, Nils Jansen, Erika {\'{A}}brah{\'{a}}m, Joost-Pieter Katoen, and
  Bernd Becker.
\newblock Minimal counterexamples for linear-time probabilistic verification.
\newblock {\em Theor. Comput. Sci.}, 549:61--100, 2014.

\bibitem{DBLP:conf/nfm/Winterer00020}
Leonore Winterer, Ralf Wimmer, Nils Jansen, and Bernd Becker.
\newblock Strengthening deterministic policies for pomdps.
\newblock In {\em {NFM}}, volume 12229 of {\em Lecture Notes in Computer
  Science}, pages 115--132. Springer, 2020.

\end{thebibliography}
